\documentclass[10pt,twocolumn,letterpaper]{article}

\usepackage{iccv}
\usepackage{times}
\usepackage{epsfig}
\usepackage{graphicx}
\usepackage{amsmath}
\usepackage{amssymb}
\usepackage{multirow}
\usepackage[dvipsnames]{xcolor}
\usepackage{floatrow}
\usepackage{subcaption}
\usepackage{booktabs}
\usepackage[linesnumbered,ruled,lined,boxed]{algorithm2e}

\usepackage[pagebackref=true,breaklinks=true,letterpaper=true,colorlinks,bookmarks=false]{hyperref}

\SetCommentSty{mycommfont}
\SetKwInput{kwHyper}{Hyperparameters}
\usepackage[capitalize]{cleveref}
\crefname{section}{Sec.}{Secs.}
\Crefname{section}{Section}{Sections}
\Crefname{table}{Table}{Tables}
\crefname{table}{Tab.}{Tabs.}

\iccvfinalcopy 

\def\iccvPaperID{9533} 

\ificcvfinal\fi

\definecolor{C0BLUE}{HTML}{1F77B4}
\definecolor{C3RED}{HTML}{D62728}

\begin{document}

\title{Frame Fusion with Vehicle Motion Prediction for 3D Object Detection}

\author{Xirui Li$^{1}$ \hspace{4mm}Feng Wang$^{2}$\hspace{4mm}Naiyan Wang$^{2}$\hspace{4mm}Chao Ma$^{1}$\\
$^{1}$Shanghai Jiao Tong University\hspace{4mm}$^{2}$Tusimple
\\
{\tt\small \{lixirui142,chaoma\}@sjtu.edu.cn \hspace{0.5mm} \{feng.wff,winsty\}@gmail.com}
\vspace{-2.5mm}
}

\maketitle
\ificcvfinal\thispagestyle{empty}\fi

\begin{abstract}
In LiDAR-based 3D detection, history point clouds contain rich temporal information helpful for future prediction. In the same way, history detections should contribute to future detections. In this paper, we propose a detection enhancement method, namely FrameFusion, which improves 3D object detection results by fusing history frames. In FrameFusion, we ``forward'' history frames to the current frame and apply weighted Non-Maximum-Suppression on dense bounding boxes to obtain a fused frame with merged boxes. To ``forward'' frames, we use vehicle motion models to estimate the future pose of the bounding boxes. However, the commonly used constant velocity model fails naturally on turning vehicles, so we explore two vehicle motion models to address this issue. On Waymo Open Dataset, our FrameFusion method consistently improves the performance of various 3D detectors by about $2$ vehicle level 2 APH with negligible latency and slightly enhances the performance of the temporal fusion method MPPNet. We also conduct extensive experiments on motion model selection.

\end{abstract}

\section{Introduction}

\begin{figure}[t]
\centering
\begin{subfigure}{0.48\linewidth}
\includegraphics[width=\linewidth]{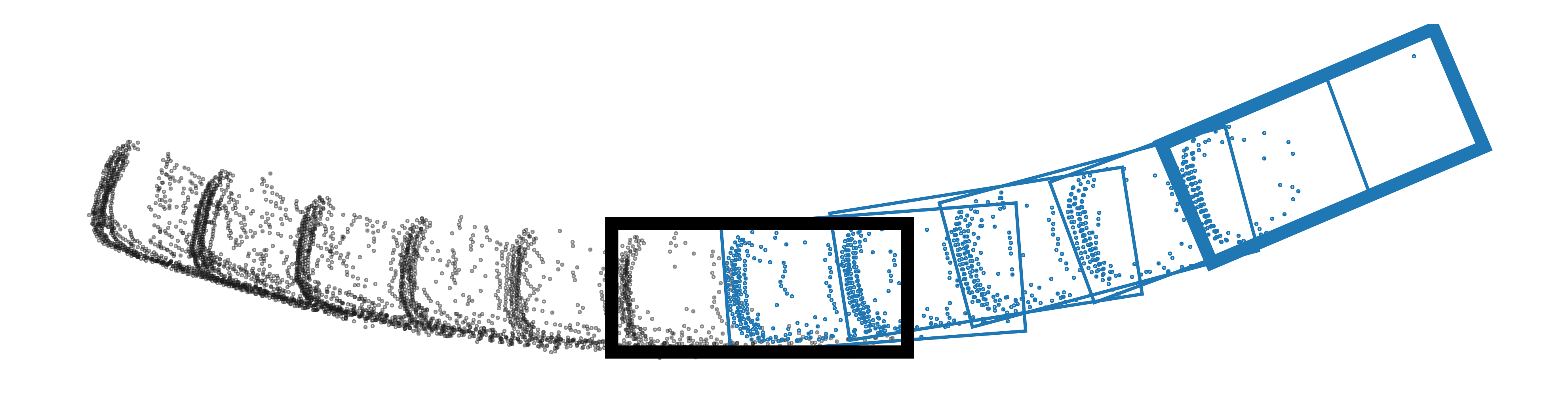}
\caption{GT trajectory}
\label{fig:intro_ill_gt}
\end{subfigure}
\hfill
\begin{subfigure}{0.48\linewidth}
\includegraphics[width=\linewidth]{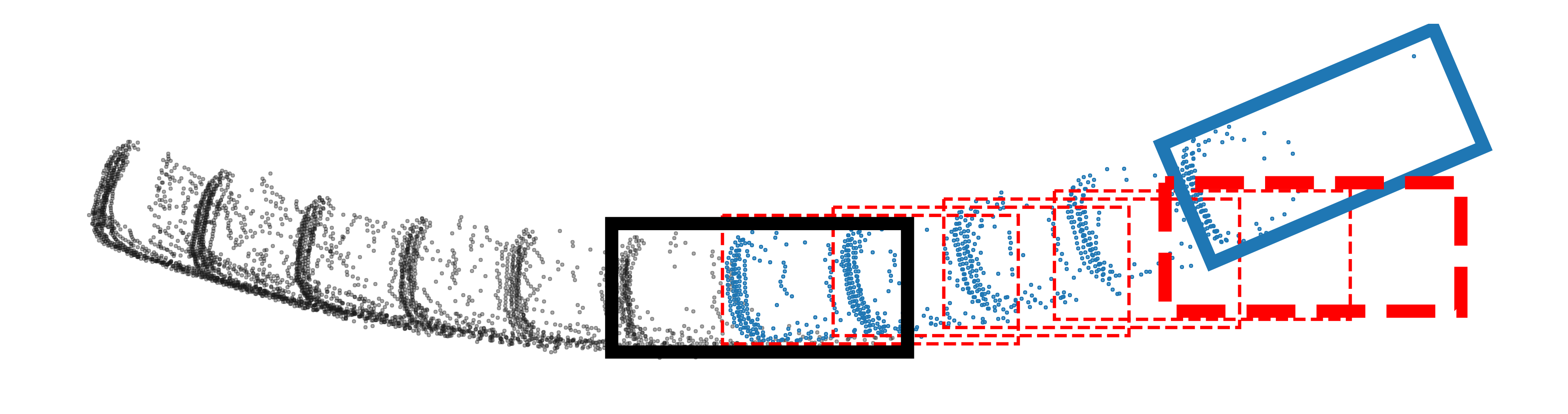}

\caption{CV motion estimation}
\label{fig:intro_ill_vxvy}
\end{subfigure}
\\
\begin{subfigure}{0.48\linewidth}
\includegraphics[width=\linewidth]{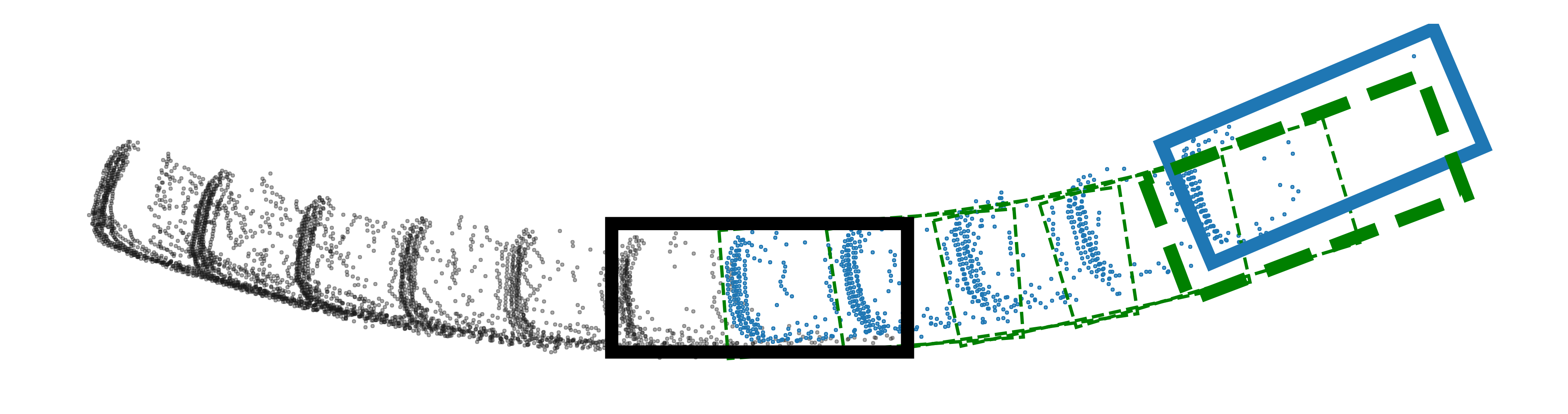}
\caption{Unicycle motion estimation}
\label{fig:intro_ill_uni}
\end{subfigure}
\hfill
\begin{subfigure}{0.48\linewidth}
\includegraphics[width=\linewidth]{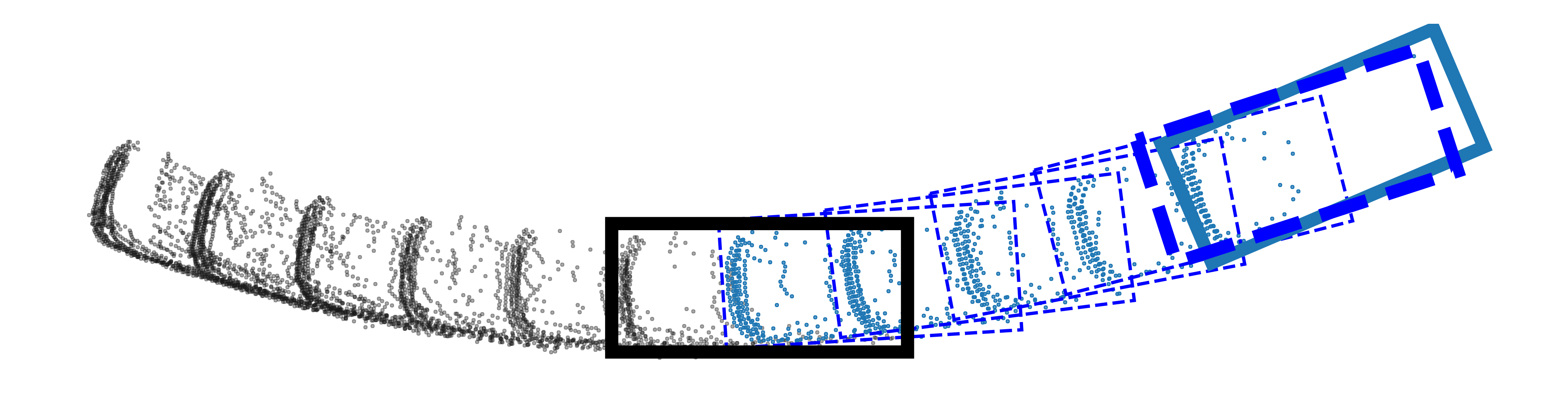}
\caption{Bicycle motion estimation}
\label{fig:intro_ill_bic}
\end{subfigure}

\caption{We propose a detection enhancement method that fuses the current detection frame with history frames leveraging motion models for pose estimation. A vehicle always moves along a trajectory in continuous frames (\textbf{current}, \textcolor{C0BLUE}{future}). We can predict its motion with the input point cloud sequence and help with future detection. The more accurate the future trajectory is estimated, the better enhancement can be expected. Constant velocity model deviates as the vehicle turns. Unicycle model turns but is still not fully aligned. Bicycle model fits the trajectory almost accurately.}
\label{fig:intro_ill}
\end{figure}

LiDAR-based 3D object detection is a crucial component of modern autonomous driving systems. To overcome the inherent sparsity of LiDAR scans, a common practice is to utilize temporal information by incorporating multiple frame point clouds as input. For example, we typically use $10$ sweeps on nuScenes dataset~\cite{caesar2020nuscenes} and $2-5$ frames on Waymo Open Dataset~\cite{sun2020scalability}. The point cloud sequences are usually transformed to the latest timestamp using accurate sensor poses. Then, all the points from multiple sweeps or frames are concatenated together and fed into the network~\cite{yin2021center, ge2020afdet}. This approach results in denser static objects and moving objects with "tails" (Figure \ref{fig:intro_ill}), which provide more prominent features for detectors.

With the temporal information, some work additionally adds a planar velocity regressor (i.e., the speed for $x$ and $y$ axes) for downstream applications such as object tracking~\cite{yin2021center} and trajectory prediction. But the regressor has not been shown to benefit the detection task itself. However, recent research in 3D object tracking~\cite{wang2021immortal} suggests that \emph{a simple motion model in 3D space is sufficient to predict the states of objects in future frames}. While this principle is difficult to apply to 2D object detection due to perspective projection, it is straightforward in 3D space. Meanwhile, in 3D detection, when an object is partially occluded, detectors usually predict inaccurate bounding boxes or miss it entirely, even if its previous observations are reliable. Our work is motivated by this insight and seeks to enhance detection performance by exploiting temporal information with motion models.

With the estimated velocities, we can construct a naive motion model to predict the trajectories of detected objects. For weakly detected objects in the current frame, we can supplement them with results forwarded by the motion model from previous frames, potentially improving the detection performance.  Importantly, our method does not explicitly rely on object tracking and is solely based on short-term prediction, making it immune to inaccuracies in tracking. By fusing detected objects from the current frame with history frames, our approach enhances detection performance, and we refer to it as \textbf{FrameFusion}.

To ensure that our method can handle more complex motion states in practical scenarios, we have delved into other vehicle motion models. As illustrated in Figure \ref{fig:intro_ill_vxvy}, the constant velocity estimator predicts inaccurate poses for turning vehicles. To address this limitation, we introduce two additional motion models from robotic dynamics: the unicycle model and the bicycle model (Fig. \ref{fig:intro_ill_uni},\ref{fig:intro_ill_bic}). These models consider the motion constraints of turning vehicles and produce more robust results for such cases, particularly for the bicycle model. It is important to note that our exploration of different motion models is not intended to improve the performance of our method on benchmarks, as turning vehicles constitute only a negligible portion of the dataset. Rather, our goal is to avoid introducing errors in turning cases that are not expected in real-world situations.

 Recently, some temporal fusion methods have shown impressive detection performance, such as MPPNet~\cite{chen2022mppnet}, which employs a carefully-designed three-level hierarchical framework to encode and fuse features from multiple frames. It adopts a first-stage 3D detector to generate 3D proposal trajectories and leverages motion information from the proposal trajectories, which is similar to our method. While MPPNet achieves very high detection performance improvement with its elaborate structure, it requires two-stage training and introduces additional latency during inference. In contrast, our method delivers a considerable enhancement with negligible latency cost as a simple postprocessing step. Moreover, we show our method can improve the detection performance based on the MPPNet result. It proves our method is complementary to the feature-level temporal fusion methods to some extent.

We evaluate our method on various existing 3D detectors and multiple 3D detection datasets. On Waymo Open Dataset, our method improves the performance of CenterPoint~\cite{yin2021center} with point cloud sequence as input by $2.1$ in level 2 vehicle APH and achieves a slight $0.6$ enhancement on MPPNet-4frame~\cite{chen2022mppnet}. Further experiments demonstrate the effectiveness of both unicycle and bicycle models for turning vehicles.

In brief, our contributions are:
\begin{itemize}
     \item We propose a detection enhancement method that improves LiDAR-based 3D object detection results by a postprocessing fusion step. Unlike past feature-level or point-level fusion methods, our method directly exploits temporal information in the detection frames by fusing boxes from previous frames forwarded with motion models.
    
    \item We first introduce vehicle motion models into the 3D object detection task. We explore the unicycle model and the bicycle model to alleviate the motion prediction error in the turning cases.
    
    \item Evaluation on benchmarks demonstrates that our method can enhance the detection performance of various 3D detection methods with negligible cost. We also conduct extensive experiments on the selection of motion models.
\end{itemize}





\section{Related Work}
\paragraph{3D LiDAR Object Detection} 3D LiDAR object detection methods can be generally divided into two categories by point cloud representation: point-based and voxel-based. Point-based methods~\cite{shi2019pointrcnn, yang20203dssd, yang2019std, qi2019deep} directly extract point features on raw irregular 3D point clouds supported by PointNet~\cite{qi2017pointnet} and then perform 3D detection. Voxel-based methods~\cite{lang2019pointpillars, yan2018second, yang2018hdnet, yang2018pixor, zhou2018voxelnet, yin2021center} instead divide the points into fixed-size voxels and apply highly efficient 3D or 2D convolutions for detection. In this work, we base most of our experiments on CenterPoint~\cite{yin2021center} while including a comparison of our method performance on multiple 3D detectors~\cite{yan2018second, shi2020pv, chen2022mppnet}.
\paragraph{Temporal Information Exploitation}  Exploiting temporal information in point cloud sequences proves to be a practical approach. A simple multi-frame point cloud concatenation adopted in some recent work~\cite{hu2022afdetv2, sun2021rsn, piergiovanni20214d, yin2021center, caesar2020nuscenes} can lead to an improvement compared to single frame input. There are a series of methods diving deeper into modeling the temporal information interaction at the feature level~\cite{yang20213d, huang2020lstm, yin2020lidar, qi2021offboard, luo2021exploring, chen2022mppnet}. 3D-Man~\cite{yang20213d} utilizes a memory bank to store temporal features and aggregate them through an attention network. Offboard3D~\cite{qi2021offboard} significantly improves the detection performance in the off-board scenario by utilizing the whole point cloud sequence. MPPNet~\cite{chen2022mppnet} recently proposed a sophisticated three-level hierarchical framework that incorporates temporal information from a long sequence using proxy points. 


Unlike the feature-level fusion methods mentioned earlier, our FrameFusion technique directly fuses temporal information at the frame level. This approach enables our method to leverage the abundant information present in correlated detection boxes across continuous frames. As a post-processing step, FrameFusion is independent of any specific detection model and is much faster than previous feature-level temporal fusion methods (as efficient as a normal NMS step). Our experiments demonstrate that our method can enhance the performance beyond the results obtained by MPPNet, suggesting that our method is complementary to other temporal fusion methods.

\begin{figure*}[ht]
  \centering
   \includegraphics[width=\linewidth]{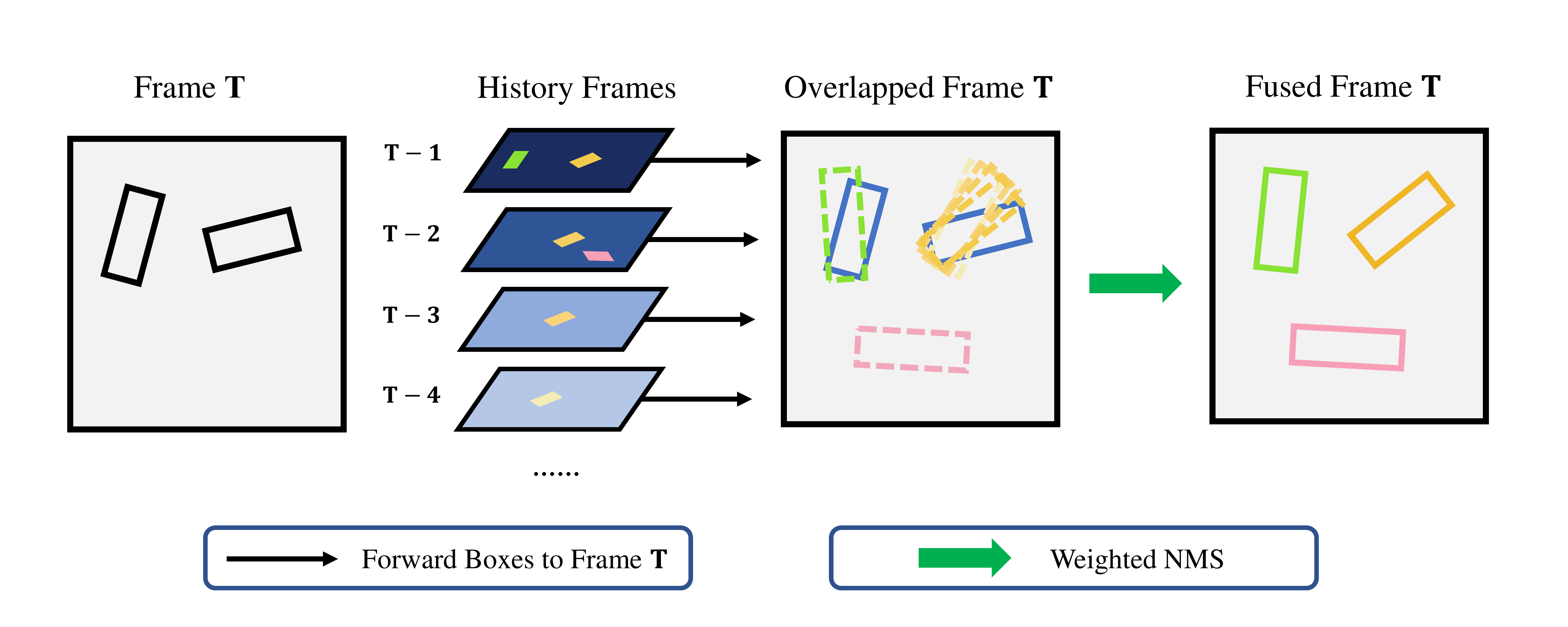}
   \caption{Frame fusion procedure. To enhance a detection frame, first select a set of history frames. Then, move forward detected boxes from history frames to the latest frame with vehicle motion models. It generates an overlapped frame with dense boxes. Apply weighted NMS on the dense boxes to get a fused frame as the frame fusion output.}
   \label{fig:frame_fusion}
\end{figure*}

\section{Preliminaries}
\paragraph{3D Detection} 
In 3D point clouds, an object is represented by a 3D bounding box $b=(x,y,z,w,l,h,\phi)$ where $(x,y,z)$ is the center location, $(w,l,h)$ is 3D size and $\phi$ is the box heading angle. Given an orderless point cloud $\mathcal{P}=\{(x,y,z)_i\}$, 3D object detection aims to predict a set of 3D object bounding boxes $\mathcal{B}=\{b_k\}$ for all the target objects in the given point cloud. We use a concatenated point cloud sequence as input in our temporal setting. To encode temporal information, we append a time indicator $t$ to the point, and thus the input point cloud is $\mathcal{P}_t=\{(x,y,z,t)_i\}$.

\paragraph{Motion Model for 3D Bounding Boxes}
Among the 3D bounding box attributes, $(w,l,h)$ is constant for a certain object. We regard the others as object pose $p(t)=\left(x(t),y(t),z(t),\phi(t)\right)$, or $p_t$ in short. It indicates the location and orientation of an object at a certain moment $t$.

We define the motion model on 3D bounding boxes as the pose derivative with respect to time. A motion model $M$ makes an assumption about the object dynamics with a set of motion parameters $\theta_{M}$, 
\begin{equation}
  M(p; \theta_{M})=\frac{\partial p}{\partial t}.
  \label{eq:motion_model}
\end{equation}

Given a motion model $M$ and the object motion parameters $\theta_{M}$, the pose at time $t$ can be estimated by integration,
\begin{equation}
  p_t=p_0 + F_{M}(p_0,t;\theta_{M}) = p_0+\int_{0}^{t}M(p_t;\theta_{M})dt,
  \label{eq:forward_model}
\end{equation}
where $p_0$ is the initial object pose. We denote the integration function $F_{M}(p_0,t;\theta_{M})$ as the forward model of motion model $M$.

In this paper, we only model vehicle objects. For vehicles, the $z$-axis variation is negligible. So we only keep the object pose as $p=(x,y,\phi)$ in the remaining sections.
\section{Method}
In this section, we first elaborate on our proposed FrameFusion algorithm, following a description of two motion models introduced for turning vehicles. We explain our model implementation at last.

\subsection{FrameFusion}
The key idea of FrameFusion is to fuse the current frame with history frames leveraging motion models for pose estimation. Fig. \ref{fig:frame_fusion} outlines the steps involved in frame fusion. For one specific frame, we forward history frame boxes to this frame with vehicle motion models. Then dense overlapped boxes are fused via weighted NMS. The use of weighted NMS ensures that dense boxes are fused rather than simply being selected, thereby facilitating the fusion of temporal information from different frames more effectively than standard NMS.

For a target frame $t$, let $\{b_{t}\}$ be the detected bounding box set. We select $N$ history frames with their detected bounding boxes set $\{b_{t-n}, \dots, b_{t-1}\}$, and then forward the history bounding boxes to frame $T$ with forward model (Equation \ref{eq:forward_model}) to get $\{ b^{for}_{t-n}, \dots, b^{for}_{t-1} \}$. Since the forwarded pose may not be that accurate, we calculate a decayed confidence score with a weight decay factor $d$, serving as the voting weight in the next step. For $i_{th}$ history frame, decayed scores are the box scores multiplied by $d^{i}$.
 

We fuse the dense detection set by applying weighted non-maximum suppression (weighted NMS)~\cite{Gidaris_2015_ICCV}. Similar to standard NMS, we first sort the bounding boxes by their scores. For each top-ranked bounding box $b$, we select a set of boxes overlapped with $b$ with IoU larger than $h_{low}$. Then bounding boxes whose IoUs with $b$ are larger than $h_{high}$ is voted together, 
\begin{equation}
        b_f=
        \frac{\sum_{b_k \in \mathcal{N}(b;h_{high})} w_k\cdot b_k}{\sum_{b_k \in \mathcal{N}(b;h_{high})} w_k},
    \label{eq:weighted_nms}
\end{equation}
where $w_k$ is the (decayed) confidence score of $b_k$ and  $\mathcal{N}(b;h_{high})=\{b_k|IoU(b,b_k)>h_{high}\}$. All the box attributes, as well as the motion parameters and the box scores, are fused as Equation \ref{eq:weighted_nms}. Scores of boxes that are only fused by history bounding boxes are reduced to avoid affecting the current frame detection. We provide additional details in the supplementary materials. Bounding boxes whose IoUs with $b$ are in the range $[h_{low}, h_{high}]$ are discarded. We iterate the above procedure until all bounding boxes are fused together as a set $\mathcal{B}_f$, which is our frame fusion output.

Overall, our FrameFusion technique can be regarded as a post-processing method that can be applied to any continuous detection frames with motion parameters prediction. In the next section, we will further introduce the motion models we used, which determine the forward model and the motion parameters in FrameFusion.

\subsection{Motion Model}

In our work, we explore three motion models for vehicle objects. It is important to note that the selection of these models is not aimed at improving benchmark performance. In fact, adopting different motion models achieves similar results on the benchmarks. Rather, the introduction of the unicycle and bicycle models is intended to alleviate the motion prediction error in turning cases, which we are concerned about in practical scenarios. We will further demonstrate these models in the following sections.

\paragraph{Constant Velocity (CV) Model}
Estimating velocities on $xy$-axes is one of the defined tasks on nuScenes Dataset, and the $xy$ velocities are also provided as ground-truth labels for Waymo Dataset. So it is a common practice to predict the $xy$ velocities if the input is multi-frame or multi-sweep. Its motion model can be simply derived as $V(p,\theta_V)=(v_x,v_y,0)$, and its forward model is $F_V(p_0,t;\theta_{V})=(v_x t,v_y t, 0)$ with $\theta_{V}=(v_x,v_y)$.

In this paper, we regard it as our baseline motion model. It performs well with our frame fusion method on benchmarks. However, it fails to describe the motion of turning vehicles. We will introduce the following two vehicle models to fix this issue.
\paragraph{Unicycle Model}
\begin{figure}[t]
  \centering
    \begin{subfigure}{0.48\linewidth}
    \includegraphics[width=\linewidth]{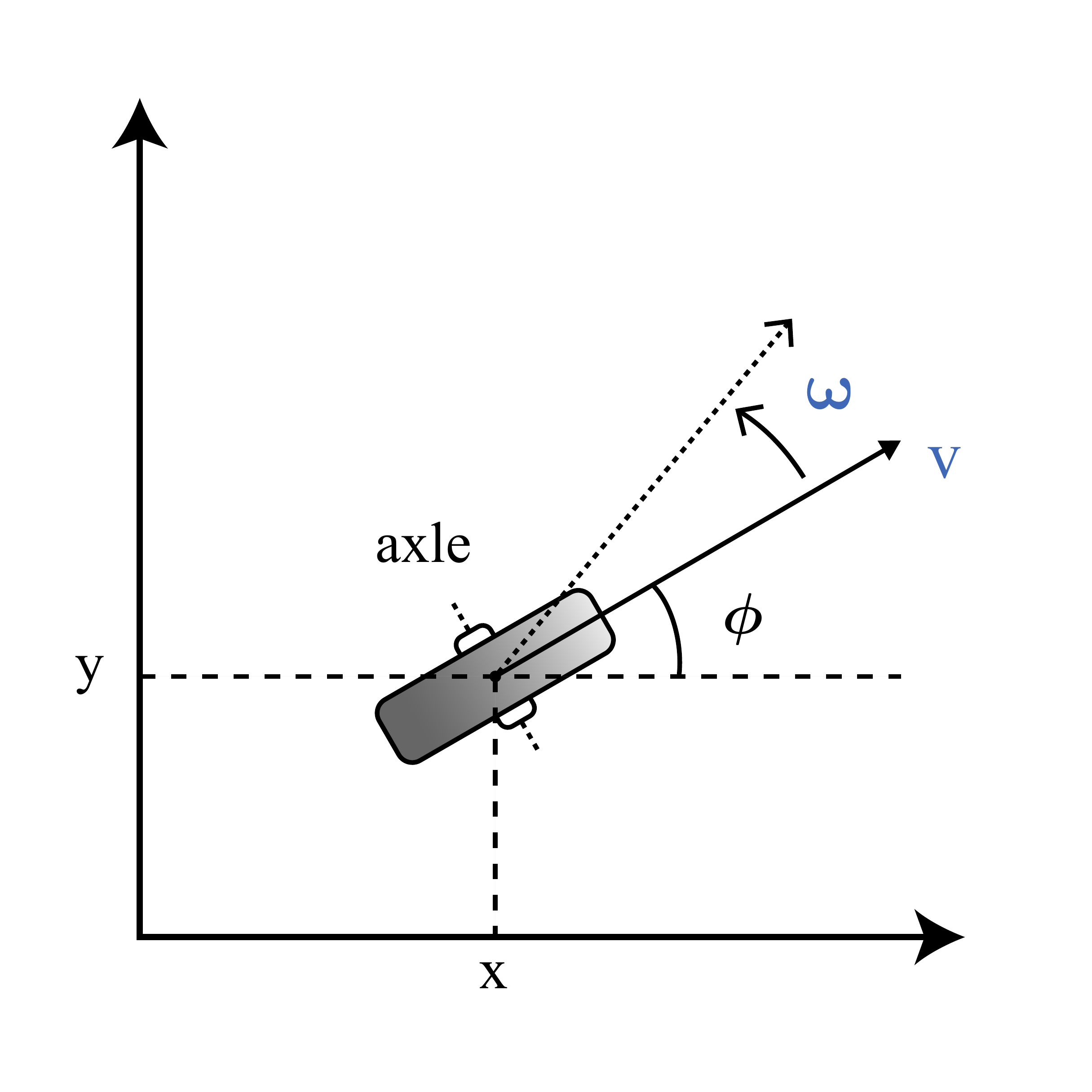}
    \caption{Unicycle model}
    \label{fig:unicycle}
  \end{subfigure}
  \begin{subfigure}{0.48\linewidth}
    \includegraphics[width=\linewidth]{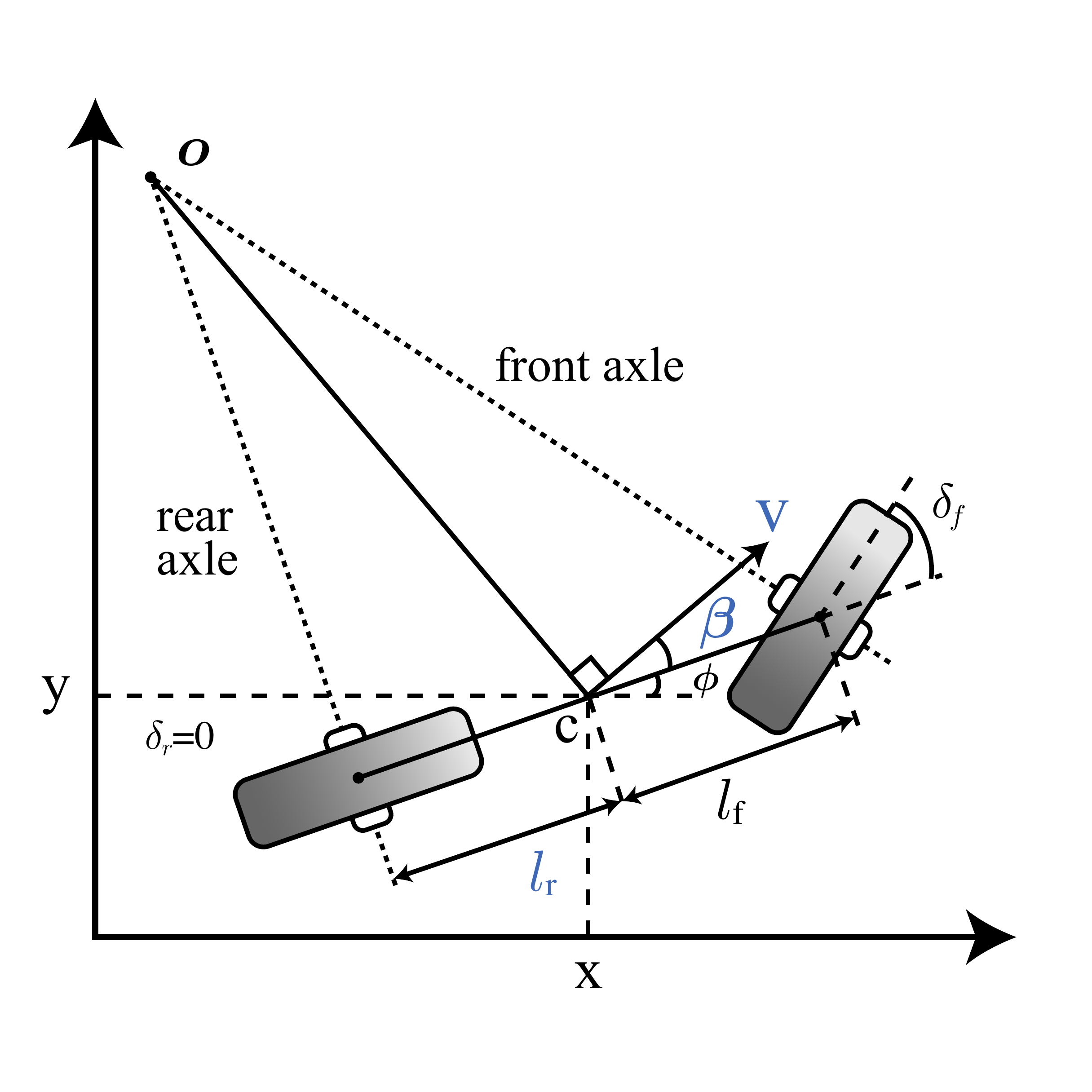}
    \caption{Bicycle model}
    \label{fig:bicycle}
  \end{subfigure}

  \caption{Unicycle model and bicycle model diagram. For unicycle model, The wheel moves with velocity $V$ and rotates with angular velocity $\omega$. Unicycle model parameter $\theta_{U}=(V,\omega)$. For bicycle model, the vehicle velocity $V$ deviates from the vehicle heading by a slip angle $\beta$. Bicycle model parameter $\theta_{\beta}=(V,\beta,l_r)$. The lighter part indicates the wheel heading. }
  \label{fig:motion_model}
\end{figure}
Unicycle model assumes a single-axle system. Suppose the vehicle is driven by a single wheel (Fig. \ref{fig:unicycle}), whose pose is $p_w=(x,y,\phi)$. It can move along the heading angle $\phi$ with speed $V$. At the same time, it can also rotate its heading with angular speed $\omega$. Unicycle model has two parameters $\theta_{U}=(V, \omega)$. Here $V$ is the signed speed. Negative $V$ means opposite to wheel heading and vice versa. A vehicle shares the same pose with the wheel, $p=p_w=(x,y,\phi)$. According to Equation \ref{eq:motion_model}, we give the unicycle model $U$,
\begin{equation}
  U(p; \theta_{U})=\frac{\partial p}{\partial t}=    
  \begin{bmatrix}
    V\cos\phi\\
    V\sin\phi\\
    \omega\\
    \end{bmatrix}.
  \label{eq:unicycle_model}
\end{equation}

By integrating over time $t$, we can estimate pose at time $t$ with the forward model $F_{U}$ (Equation \ref{eq:forward_model}),
\begin{equation}
    \label{eq:unicycle_forward}
    p_t=p_0+F_{U}(p_0,t;\theta_{U})=p_0+
    \begin{bmatrix}
    \frac{V}{\omega}(\sin\phi_t-\sin\phi_0)\\
    \frac{V}{\omega}(\cos\phi_0-\cos\phi_t)\\
    \omega t\\
    \end{bmatrix},
\end{equation}
where $\phi_{t}=\phi_{0}+\omega t$ is the heading at time t.


The unicycle model provides a compact way to describe a rigid motion with both heading and position variation. However, it is not accurate enough for most vehicles, which usually have two axles instead of a single one. Nonetheless, because of its simplicity, it is also widely used as an approximation in mobile robotics~\cite{lee2001tracking,ghommam2010formation}. 
\paragraph{Bicycle Model}
Bicycle model represents a system with two axles separated by a distance (Fig. \ref{fig:bicycle}). The vehicle is driven by two wheels distributed on the vehicle center line. For simplicity, we assume the rear wheel is parallel with the vehicle heading $\phi$. The front wheel is the turning wheel with a rotation angle difference $\delta_f$ over $\phi$. The distances between the center of mass $C$ and two wheels are $l_f,l_r$.

In the bicycle model, the vehicle pose is not identical to the wheel as in the unicycle model. Thus the velocity does not always follow the vehicle heading angle $\phi$. With a nonzero rotation of the front wheel, the front axle and the rear axle intersect at a point $O$. The velocity $V$ is orthogonal to line $OC$. We define the angle between vehicle heading and velocity direction as the tire slip angle $\beta$. $\beta$ is $\pi$-normalized to $[-\pi / 2, \pi / 2]$.

We derive the bicycle model as
\begin{equation}
  B(p; \theta_{B})=\frac{\partial p}{\partial t}=    
    \begin{bmatrix}
    V\cos(\phi+\beta)\\
    V\sin(\phi+\beta)\\
    \frac{V\sin\beta}{l_r}
    \end{bmatrix}.
  \label{eq:bicycle_model}
\end{equation}
The bicycle model has three parameters $\theta_{B}=(V,\beta,l_r)$. $\beta$ depends on the front wheel rotation angle $\delta_f$,
\begin{equation}
\label{eq:bicycle_beta}
    \beta=\arctan(\frac{l_r}{l_f+l_r}\tan\delta_f).
\end{equation}
Note that $\beta,\delta_{f}$ are equivalent in the bicycle model with given $l_f$ and $l_r$. We choose $\beta$ as a motion parameter for compactness.

Similar to the unicycle model, we have the bicycle forward model by Equation \ref{eq:forward_model},
\begin{equation}
    \label{eq:bicycle_forward}
    p_t=p_0+
    \begin{bmatrix}
    \frac{l_r}{\sin\beta}(\sin(\phi_t+\beta)-\sin(\phi_0+\beta))\\
    \frac{l_r}{\sin\beta}(\cos(\phi_0+\beta)-\cos(\phi_t+\beta))\\
    \frac{V\sin\beta}{l_r} t
    \end{bmatrix},
\end{equation}
where $\phi_{t}=\phi_{0}+\frac{V\sin\beta}{l_r} t$ is the heading at time $t$.


Compared with the more advanced Ackermann model~\cite{mueller2019modern}, the bicycle model trades off between accuracy and simplicity. We will show in experiments that the bicycle model is accurate enough to model the turning vehicles.
\paragraph{Discussion} \label{para:discussion}

In this section, we briefly discuss the pros and cons of the three aforementioned motion models. 
A naive observation is that these motion models are the same when the vehicle moves straight. We can set $\omega$ and $\beta$ in the unicycle and bicycle model to zero and get the same forward model as the constant velocity model. So we only discuss the turning cases when comparing the motion models. 

\begin{figure}
  \centering
    \includegraphics[width=\linewidth]{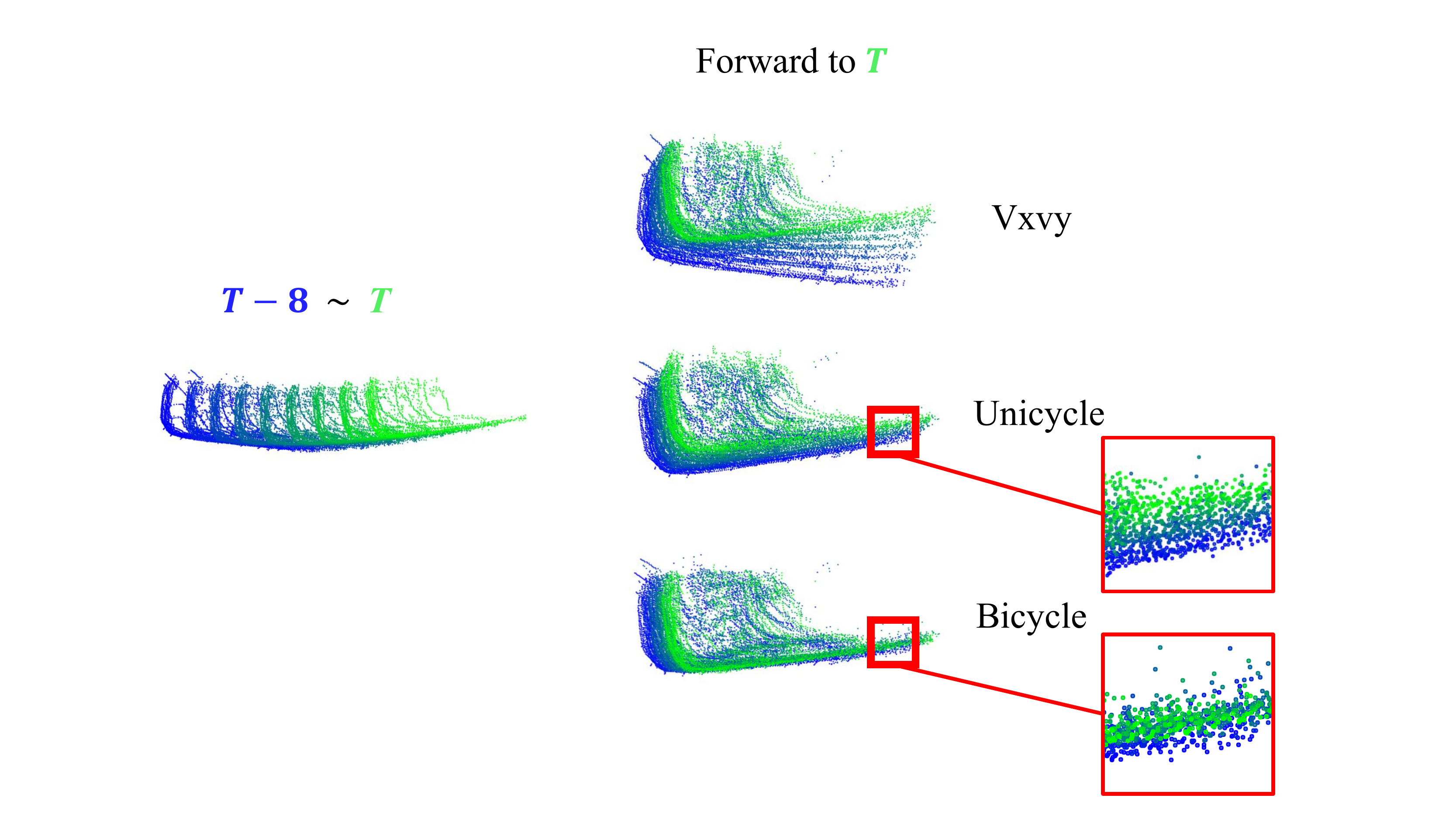}
  \caption{Forwarding history point cloud into the current frame with motion model.}
  \label{fig:discussion}
\end{figure}

Fig. \ref{fig:discussion} shows a turning vehicle point cloud sequence in eight frames. We estimate its motion parameter at each timestamp with adjacent box poses
and forward all the point clouds to the latest frame $T$. We can see the constant velocity model diverges in the heading orientation as it estimates incorrect velocity directions. In comparison, the unicycle model shows significantly better motion estimation performance with a better-aligned point cloud. However, it still shifts a little that can be figured out at the border. The bicycle model generates an almost perfect matching. We can hardly tell points from different frames, presenting its high accuracy in vehicle motion prediction.

Though the bicycle model result is satisfying, we observe a short blue tail at the rear. It actually suggests an acceleration in a short period. As we do not introduce second-order derivatives such as acceleration in our motion models for model simplicity, it may cause inevitable errors in long-term motion estimation. It also indicates the rationality behind the confidence decay mechanism in our frame fusion method.

\subsection{Model Implementation}
To extract motion information from point cloud sequences, we augment existing 3D detectors with additional motion parameter regressors. For any models mentioned in our work, multi-frame input fusion~\cite{yan2018second,piergiovanni20214d,caesar2020nuscenes} is adopted to integrate temporal information into model input for motion parameter prediction.

In the Waymo Open Dataset, we observe that vehicles with nonzero angular velocity account for only $3\%$ of the total, and their distribution is skewed around zero. When regressing angular motion parameters, this unbalanced distribution makes the network more likely to predict zero values for any target using a normal L1 loss. To address this issue, we employ a balanced loss function that equalizes the influence of turning and non-turning objects. We divide objects into two classes based on an angular velocity threshold of $0.1$: turning and non-turning. When calculating the loss of angular parameters, we first average the loss in each class and then sum them up to obtain the regression loss.

Before training, we pre-process the datasets to generate ground-truth motion parameter labels with inverse motion models. Given a pair of poses $(p_0, p_{t})$, an inverse model estimates the motion parameter $\hat{\theta_M}$. We refer the readers to supplementary material for inverse model details. For frame $t$, its motion parameters are estimated using object poses in adjacent frames $t-1$ and $t+1$ as $(p_0, p_{t})$.

\section{Experiments}
\subsection{Experimental Settings}
\paragraph{Waymo Open Dataset}
Waymo Open Dataset~\cite{sun2020scalability} is a large-scale, diverse 3D detection dataset, including three classes: vehicle, bicycle, and pedestrian. The ofﬁcial 3D detection evaluation metrics are standard 3D bounding box mean average precision (mAP)~\cite{everingham2010pascal} and mAP weighted by heading accuracy (mAPH). As we solely consider vehicle class in evaluation, tables show AP and APH on vehicle class.
The annotated objects are split into two difficulty levels: LEVEL 1 (More than ﬁve Lidar points in bbox), and LEVEL 2 (at least one Lidar point in bbox). 

\paragraph{nuScenes}
nuScenes~\cite{caesar2020nuscenes} comprises 1000 scenes, with the interval between adjacent annotated frames being 0.5s (0.1s in Waymo). In our experiments, we only consider 5 vehicle classes in the 10 providing object classes. The primary 3D detection metrics are mean Average Precision (mAP) and nuScenes detection score (NDS) ~\cite{caesar2020nuscenes}. 


\paragraph{Implementation Details}
In experiment sections, initially capitalized vehicle motion models represent our models built on the CenterPoint-Voxel model with corresponding motion parameter regressors such as Constant Velocity (CV), Unicycle, and Bicycle model. We implement these CenterPoint-based models on a 3D detection codebase MMDetection3D ~\cite{contributors2020mmdetection3d}. We implement the various 3D detectors in Table \ref{tab:large_comp} on another open 3D detection codebase OpenPCDet~\cite{openpcdet2020} and adopt the constant velocity model for these methods.

All models use multi-frame point fusion as input ($[-4,0]$ for Waymo and $[-10,0]$ for nuScenes). We train CenterPoint-based motion models with AdamW optimizer and a cyclic learning rate scheduler with a maximum learning rate 1e-3 and a weight decay $0.01$. Train for 12 epochs on Waymo and 20 epochs on nuScenes. Models in Table \ref{tab:large_comp} are trained with the default configuration.


For frame fusion, we set the frame fusion range $N=4$ by default (fuse 4 previous frames). Configurations of other parameters are left in supplementary materials.

\subsection{Main Results}
\begin{table}\small
  \centering
      \begin{tabular}{@{}lcccc@{}}
    \toprule
    \multirow{2}{*}{Method} & \multicolumn{2}{c}{AP/APH$\uparrow$ (VEH)} & \multirow{2}{*}{\#Params$^{*}$} & \multirow{2}{*}{Latency$^{*}$}\\
\cmidrule(lr){2-3}
      & Level 1 & Level 2\\
        \midrule 
    SECOND~\cite{yan2018second} &  71.9/71.2 & 63.9/63.3 & \ \ 5.3M& \ \ 137ms \\
    CenterPoint~\cite{yin2021center} & 76.7/76.1 & 69.1/68.6 &\ \ 7.8M & \ \ \ \ 74ms\\
    PV-RCNN~\cite{shi2020pv} & 78.1/77.6 & 70.1/69.6 & 13.5M & 1085ms\\
    MPPNet-4F~\cite{chen2022mppnet} &     81.5/81.0 & 74.0/73.6 & +7.9M & +301ms\\

    \midrule
    SECOND+ & 73.6/73.0& 65.6/65.1 & - &\ \ \ +3ms\\
    CenterPoint+ & 78.7/78.2 & 71.2/70.7 & - & \ \ \ +3ms\\
    PV-RCNN+ & 80.0/79.4& 72.1/71.6& - & \ \ \ +3ms\\
    MPPNet-4F+ &     82.0/81.6 & 74.6/74.2 & - & \ \ \ +3ms\\
    \bottomrule
\end{tabular}
  \caption{Vehicle detection enhancement over various 3D detectors on Waymo validation set. + means performance after applying frame fusion. All methods use \textbf{4-frame} input and constant velocity motion model. Latency is measured on a Nvidia 3090 GPU in milliseconds per frame. $*$ : For methods relying on initial proposals generated from other 3D detectors such as our frame fusion and MPPNet, we report the additional parameter number and latency introduced by themselves for a clear comparison.}
  \label{tab:large_comp}
\end{table}

\begin{figure*}
  \centering

  \begin{subfigure}{0.15\linewidth}
    \includegraphics[width=\linewidth]{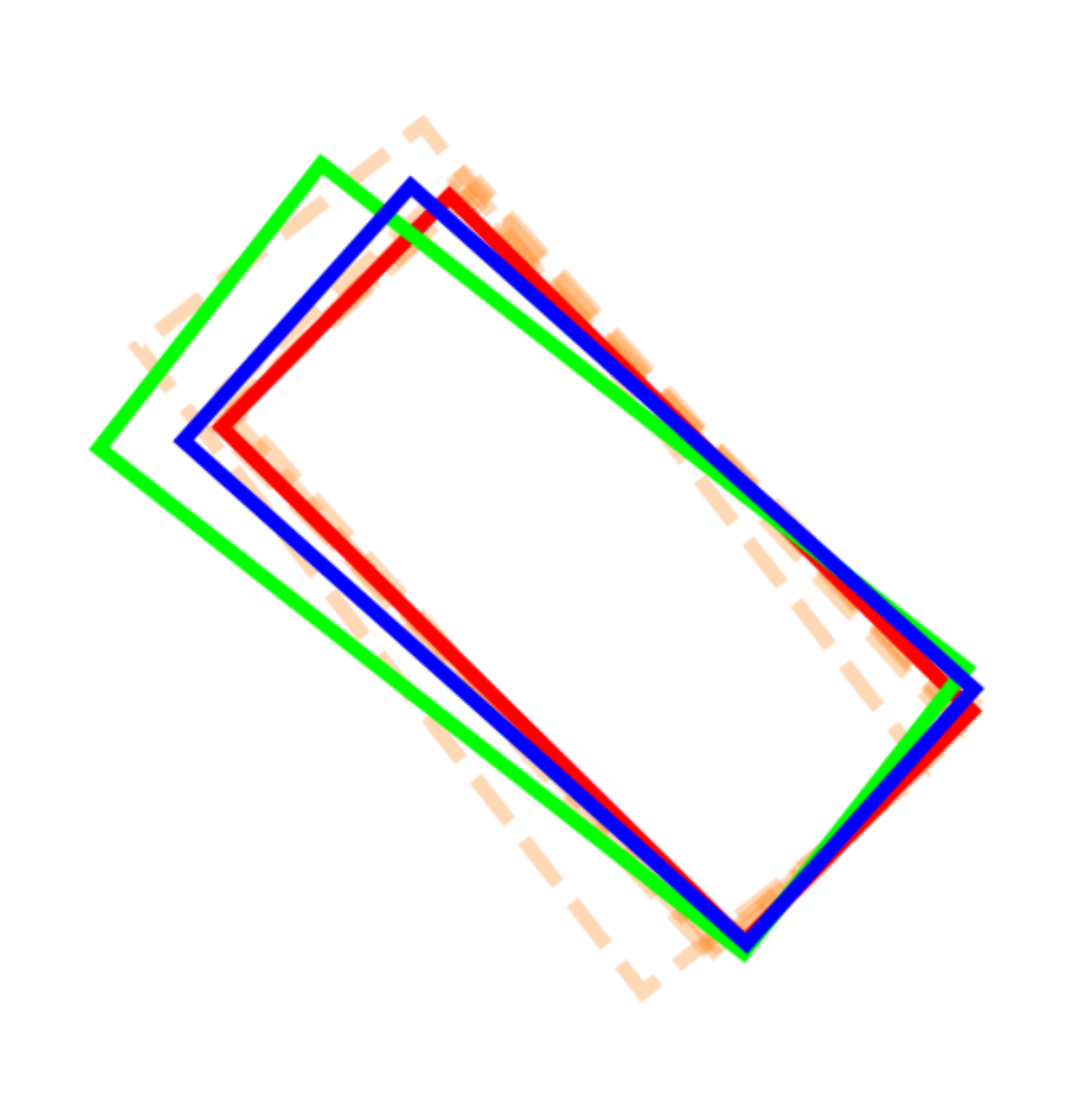}
    \caption{Box refinement}
    \label{fig:visa}
  \end{subfigure}
    \begin{subfigure}{0.15\linewidth}
    \includegraphics[width=\linewidth]{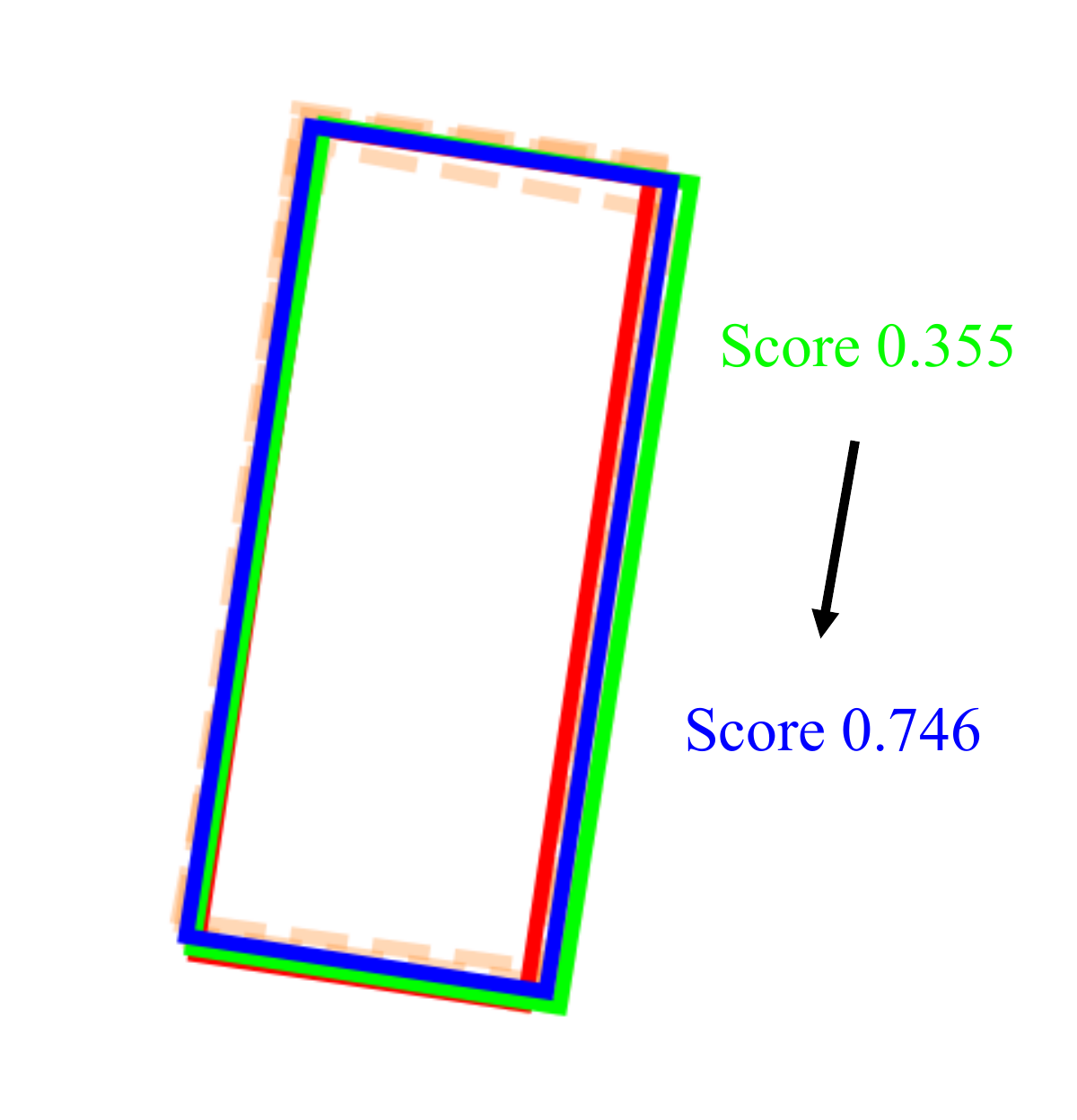}
    \caption{Score refinement}
    \label{fig:visb}
  \end{subfigure}
\begin{subfigure}{0.15\linewidth}
\includegraphics[width=\linewidth]{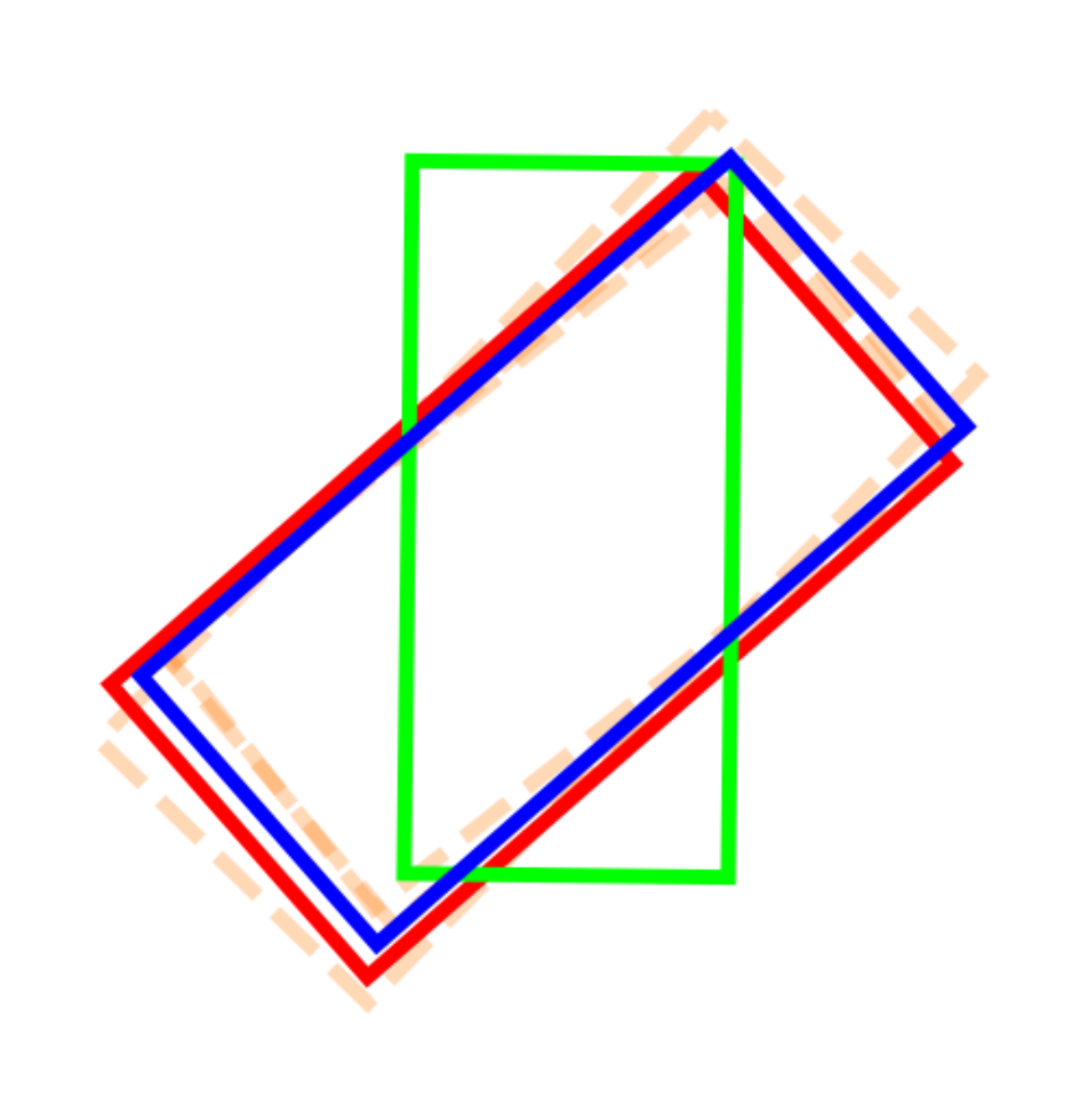}
\caption{Box covering}
\label{fig:visc}
\end{subfigure}
  \begin{subfigure}{0.15\linewidth}
    \includegraphics[width=\linewidth]{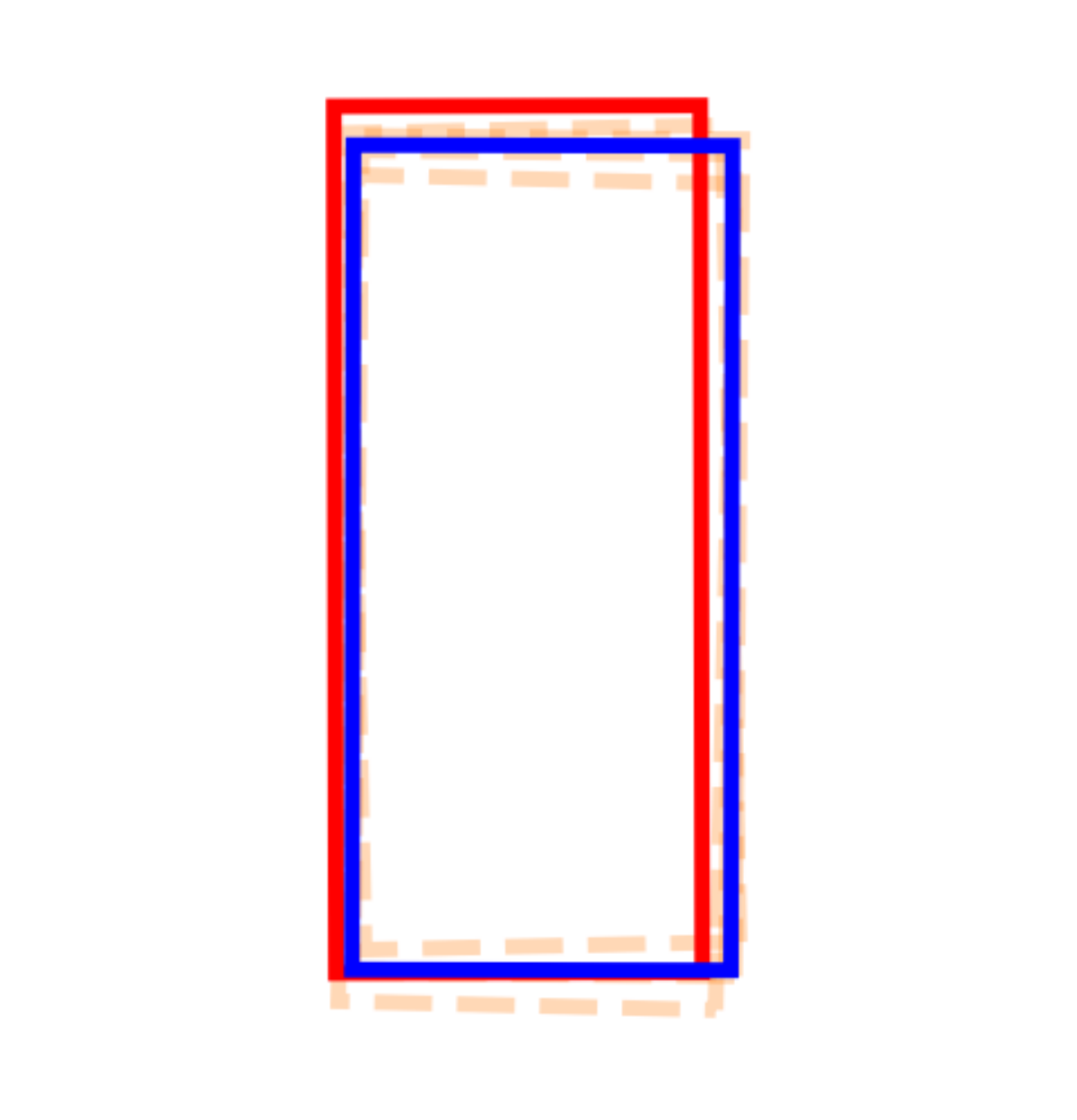}
    \caption{FN covering}
    \label{fig:visd}
  \end{subfigure}
\begin{subfigure}{0.15\linewidth}
    \includegraphics[width=\linewidth]{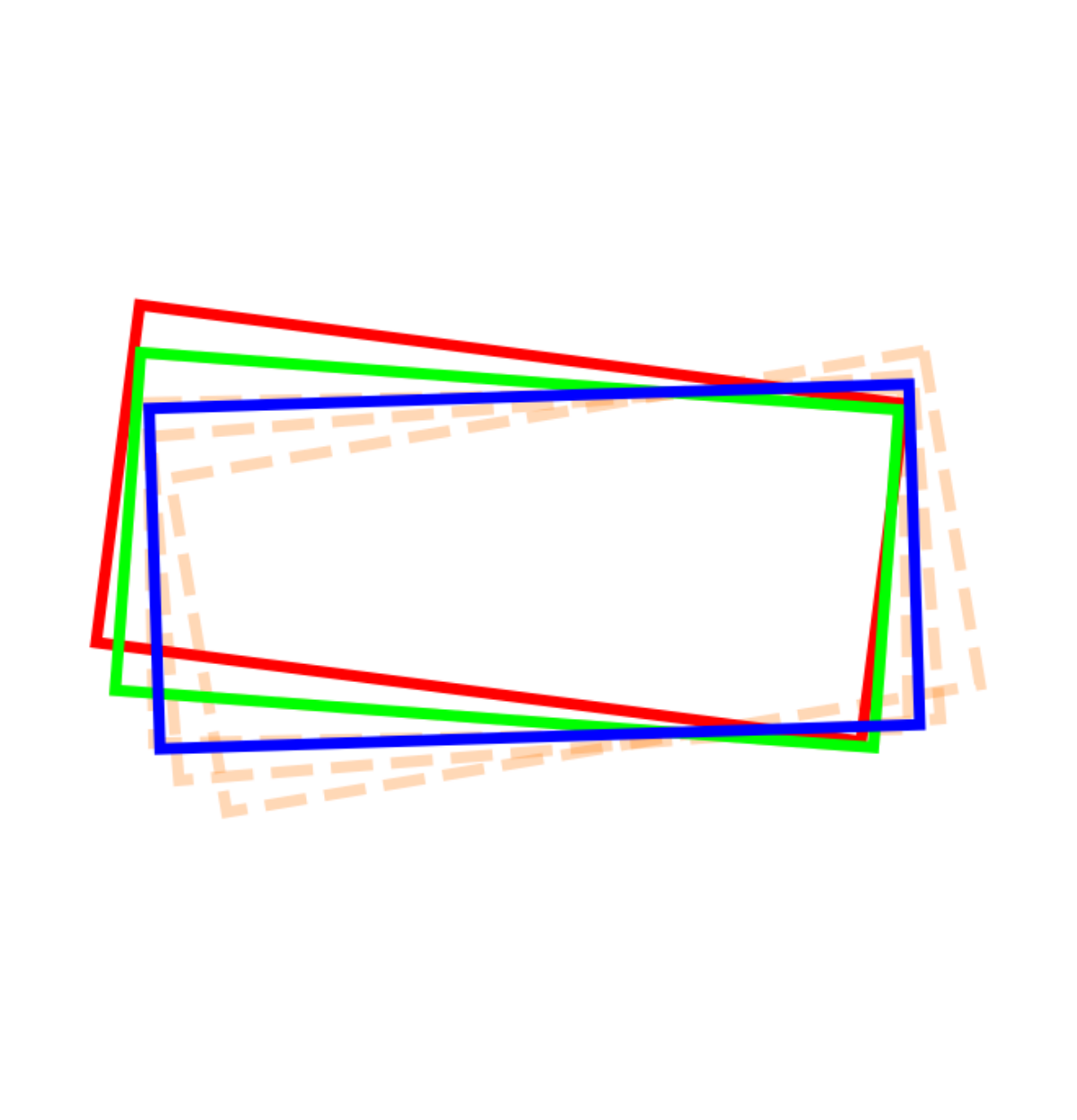}
    \caption{CV Negative}
    \label{fig:vise}
  \end{subfigure}
  \begin{subfigure}{0.15\linewidth}
    \includegraphics[width=\linewidth]{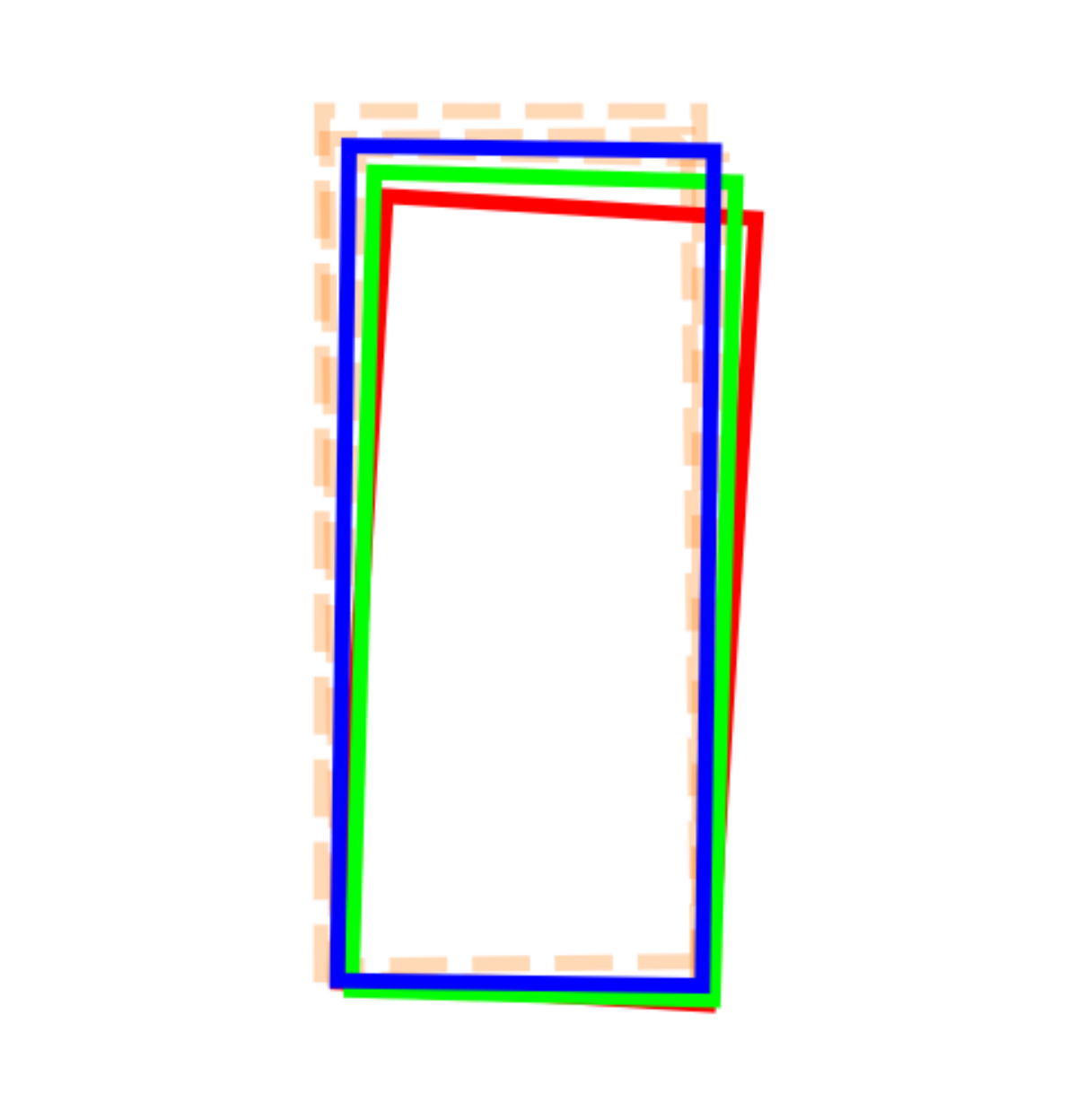}
    \caption{Bicycle Negative}
    \label{fig:visf}
  \end{subfigure}
  \caption{FrameFusion visualization on bounding boxes. Each figure depicts a \textcolor{red}{red} GT box, a \textcolor{green}{green} box from the frame before fusion and a \textcolor{blue}{blue} box from the enhanced frame after fusion. \textcolor{orange}{Orange} boxes are the estimations dense estimations before fused to the blue box.}
  \label{fig:vis}
\end{figure*}
\paragraph{3D Detection Enhancement}
\begin{table}\small
  \centering
      \begin{tabular}{lcccc}
    \toprule
    \multirow{2}{*}{Method} & \multicolumn{2}{c}{VEH AP$\uparrow$} & \multicolumn{2}{c}{VEH APH$\uparrow$}\\
\cmidrule(lr){2-3}\cmidrule(lr){4-5}
      & Level 1 & Level 2 & Level 1 & Level 2 \\
        \midrule
    CenterPoint & 76.08 &	68.17	&75.55&	67.68\\
    \midrule
    Constant Velocity &75.64&	67.74&	75.11&	67.26 \\
    Unicycle &75.38&	67.46&	74.84&	66.98\\
    Bicycle&75.56&	67.62&	75.02&	67.13\\
    \midrule
    Constant Velocity+ &77.82&	70.14&	77.33&	69.68\\
    Unicycle+ & 77.65 &69.93 &77.14 &69.46  \\
    Bicycle+ &77.84&	70.11&	77.33&	69.65 \\
    \bottomrule
\end{tabular}
  \caption{3D vehicle detection improvements on Waymo validation set with different motion models. + means performance after applying frame fusion. All models use \textbf{4-frame} input. CenterPoint~\cite{yin2021center}: backbone model without motion parameter regression.}
  \label{tab:waymo_det}
\end{table}


FrameFusion improves the 3D detection performance by fusing history detection frames with the current frame. On the Waymo dataset, we demonstrate that our method can enhance the performance of existing 3D detection methods by applying a 4-frame fusion to their results (Table \ref{tab:large_comp}). When applied to the 3D detectors SECOND~\cite{yan2018second}, CenterPoint~\cite{yin2021center}, and PV-RCNN~\cite{shi2020pv}, FrameFusion achieves significant enhancements in the vehicle 3D APH (level 2) by $1.8, 2.1, 2.0$ with negligible latency costs ($3$ms). To ensure a fair comparison with MPPNet, we use its 4-frame version, which aligns with our 4-frame fusion setting, and take CenterPoint evaluated in the same table as its first-stage model. Compared to MPPNet, our method achieves a $42\%$ $(2.1/5.0)$ enhancement over the same CenterPoint model without the retraining step, model parameters, or high latency introduced by MPPNet. Notably, our method can even improve MPPNet's performance slightly by $0.6$ (MPPNet-4F+), suggesting that our method is complementary to other temporal fusion methods to some extent.

We compare the frame fusion performance for 3D detection enhancement with different motion models based on the CenterPoint~\cite{yin2021center} backbone. On the Waymo validation set (Table \ref{tab:waymo_det}), frame fusion shows similar enhancement for Constant Velocity, Unicycle, and Bicycle model, with an increase of $+2.42, +2.48, +2.52$ in vehicle level 2 APH. Similar results are obtained on the nuScenes Dataset (Table \ref{tab:nus_det}). It indicates that FrameFusion works consistently with different motion models. In the next section, we will analyze the differences between these models in turning cases.


\begin{table} \small
  \centering
      \begin{tabular}{lcc}
    \toprule
    Method & mAP$\uparrow$ & NDS$\uparrow$\\
        \midrule
    CenterPoint & 53.44 & -\\
    \midrule
    Constant Velocity &53.67 &		61.73\\
    Unicycle &53.61	 &	61.80\\
    Bicycle &53.55	 &	61.72\\
    \midrule
    Constant Velocity+ & 55.17	&	62.78\\
    Unicycle+ & 54.87	&	62.72\\
    Bicycle+ & 55.05	&	62.77\\
    \bottomrule
\end{tabular}
  \caption{3D detection improvements on vehicle classes of nuScenes validation set with different motion models. All models use \textbf{10-sweep} input. CenterPoint~\cite{yin2021center}: backbone model without motion parameter regression.}
  \label{tab:nus_det}
\end{table}

\paragraph{Turning Case Evaluation} \label{section:turn_case}
As discussed in Sec. \ref{para:discussion}, motion models perform identically when the angular velocity is zero. In Waymo, the ratio of turning vehicles is about $0.05$. Thus previous results do not fully reflect the difference among motion models. We further explore the motion model features in the turning case.

We evaluate motion model performance on a turning vehicles subset of Waymo validation set. Considering a standard turning vehicle, we select vehicles in GT with velocity $V > 5$ and turning radius $R < 25$ as the turning subset. The detection results are matched to the turning subset to filter out non-turning vehicles. 

Table \ref{tab:waymo_det_turn} shows the evaluation results on the turning subset. Constant Velocity model decreases the performance on turning vehicles (about -3.78 level 2 APH) after frame fusion, while Unicycle model and Bicycle model still get improvement. Bicycle model shows its superiority with the best performance on the turning case.

\begin{table} \small
  \centering
      \begin{tabular}{lcccc}
    \toprule
    \multirow{2}{*}{Method} & \multicolumn{2}{c}{VEH AP$\uparrow$} & \multicolumn{2}{c}{VEH APH$\uparrow$}\\
\cmidrule(lr){2-3}\cmidrule(lr){4-5}
      & Level 1 & Level 2 & Level 1 & Level 2 \\
    \midrule
    Constant Velocity & 81.36	&80.67&	80.80&	80.12\\
    Unicycle & 80.85&	80.35&	80.17&	79.67\\
    Bicycle & 81.62&	80.93&	81.13&	80.44\\
    \midrule
    Constant Velocity+ & 78.48&	77.74&	77.07&	76.34\\
    Unicycle+ &82.39&	81.67&	81.85&	81.14\\
    Bicycle+ & \textbf{82.96}&	\textbf{82.29}&	\textbf{82.34}&	\textbf{81.67}\\
    \bottomrule
\end{tabular}
  \caption{3D vehicle detection improvements evaluated on turning subset of Waymo validation set with different motion models.}
  \label{tab:waymo_det_turn}
\end{table}

The preceding experiments confirm the effectiveness of our motion models in capturing turning targets. While any of the three models can produce similar performance improvements in overall performance, if we aim to consistently improve detection across various motion states, which is crucial in practical scenarios, the unicycle or bicycle model should be prioritized due to its superior precision in vehicle modeling.

\subsection{Ablation Studies}

\paragraph{Qualitative Analysis}\label{sec:vis}

We intuitively present our method results with a visualization in Fig. \ref{fig:vis}. Our method enhances the detection in two forms, divided by whether a current frame box is fused in the weighted NMS (refinement in Fig \ref{fig:visa}, \ref{fig:visb}) or not (covering in Fig \ref{fig:visc}, \ref{fig:visd}).  

For a slightly deviated detection (Fig. \ref{fig:visa}), the fusion box directly refines on the deviated box by merging matched boxes in the overlapped frame. High confidence detection can also enhance the matched future detection scores, filtering better proposals (Fig. \ref{fig:visb}). 

For another type, covering, the vehicle is often weakly detected in the current frame and thus results in a significant prediction error (Fig. \ref{fig:visc}) or a missing (Fig. \ref{fig:visd}). In this case, the boxes from history frames merge by themselves and fuse to generate a TP detection. 


Fig. \ref{fig:vise} shows a negative optimized turning vehicle sampled from the frame fusion result of Constant Velocity model. The forwarded box orientations deviated from the ground truth, causing the box heading to rotate wrongly after fusion. We put another Bicycle model negative sample in Fig. \ref{fig:visf}. However, the negative correction is supposed to be negligible with the confidence decay and IoU threshold in the frame fusion.

\paragraph{Analysis of Fusion Frame Number}
\begin{figure}
  \centering
   \includegraphics[width=\linewidth]{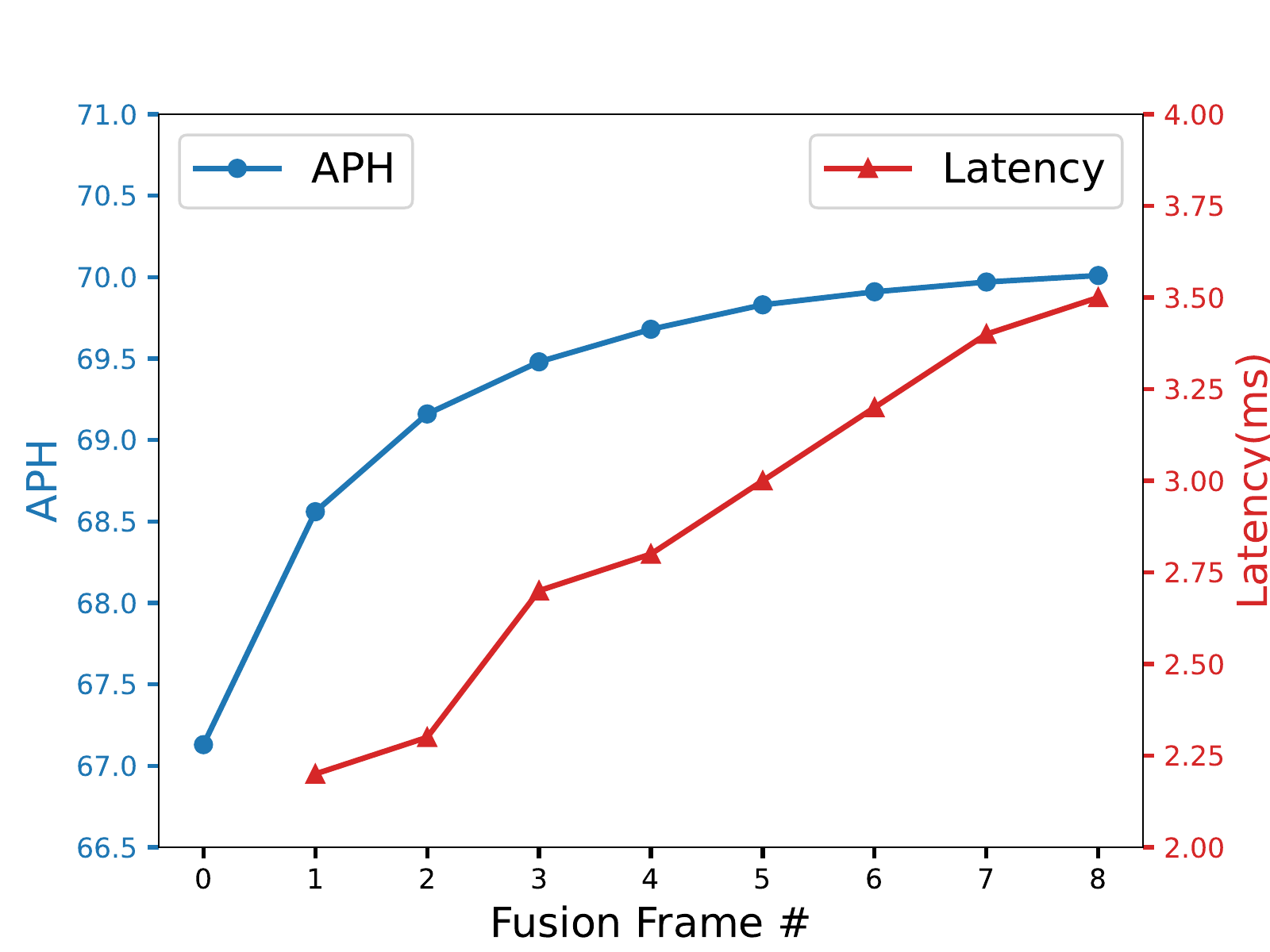}
   \caption{Detection Performance with various numbers of fusion frames evaluated with CenterPoint-Bicycle model. \textcolor{C0BLUE}{Blue}: Vehicle level 2 APH on Waymo Validation set. \textcolor{C3RED}{Red}: Latency in milliseconds per frame.}
   \label{fig:fusion_num}
\end{figure}
We analyze the performance and efficiency of our method concerning the fused frame number (Fig. \ref{fig:fusion_num}). As more frames are fused, the performance increment soon converges, while the latency increases linearly. This can be attributed to the fact that, as the fusion number grows, less additional temporal information is integrated from distant frames. Furthermore, as the time interval grows, the motion prediction becomes less accurate, and the confidence decay strategy causes the voting weight of forwarded bounding boxes to decay rapidly. Therefore, fusing a distant frame may hardly enhance current detection. In practical scenarios, fusing 2 to 4 frames can provide sufficient improvement.


\paragraph{Analysis of Vehicle Motion States}


To further analyze the performance of the motion models on different motion states, we evaluated the frame fusion enhancements separately for stationary, straight, and turning motion vehicles (Table \ref{tab:ana_mot_sta}). We found that the motion models performed almost identically when vehicles were stationary or moving straight, indicating that they were equally effective in these scenarios. However, we observed a decreasing trend in performance from stationary to straight to turning motion, which suggests a decrease in the capability of the motion models and the accuracy of the motion parameter predictions as the motion becomes more complex. The turning cases are discussed in more detail in section \ref{section:turn_case}.


\begin{table} \small
  \centering
      \begin{tabular}{lcccc}
    \toprule
    Method & All & Stationary& Straight& Turning\\
    \midrule
    Constant Velocity &        +2.42&	+2.95&	+1.24	&-3.40\\
    Unicycle & +2.48&	+2.95&	+1.39	&+0.83\\
    Bicycle & +2.52 &+3.02	&+1.20	&+1.28\\

    \bottomrule
\end{tabular}
  \caption{Compare frame fusion enhancement on different motion states. Show vehicle level 2 APH detection increment after frame fusion. Stationary, straight, turning vehicles account for $0.63$, $0.31$, $0.05$ in Waymo validation set.}
  \label{tab:ana_mot_sta}
\end{table}
\paragraph{Frame Fusion Ablation}
\begin{table} \small
  \centering
      \begin{tabular}{lcc}
    \toprule
      Method      &  APH($\Delta$ APH) & Latency\\
\midrule
CenterPoint     &   67.1    &   -       \\
CenterPoint+(Default) 	&69.7(+2.6) & 2.8ms\\
\midrule
Circle NMS &	68.2(+1.1) & 3.0ms\\
Box Refinement (Fig. \ref{fig:visa}) &	68.6(+1.5) & 2.8ms\\
Score Refinement (Fig. \ref{fig:visb}) &	67.7(+0.6) & 2.8ms\\
Only Covering (Fig. \ref{fig:visc} \ref{fig:visd}) & 67.7(+0.6) & 2.8ms\\
Only Refinement (Fig. \ref{fig:visa} \ref{fig:visb}) & 69.1(+2.0) & 2.8ms\\
    \bottomrule
\end{tabular}
  \caption{Frame fusion ablation on CenterPoint-Bicycle model. Show vehicle level 2 APH on Waymo validation set. Above: Default frame fusion result. Below: Ablation settings. We annotate some settings with their visualizations in qualitative analysis.}
  \label{tab:ff_ablation}
\end{table}
We present the results of frame fusion ablations in Table \ref{tab:ff_ablation}. Replacing weighted NMS with normal circle NMS means the new boxes are selected instead of fused, thus leading to a large performance drop. Subsequent ablations aim to quantify the contribution of each enhancement component mentioned in qualitative analysis. The latency analysis shows that our method is as efficient as a GPU-implemented NMS postprocessing step.

\section{Conclusion}
In this paper, we propose a 3D object detection enhancement method FrameFusion, fusing a detection frame with a series of history frames forwarded by motion models. We introduce vehicle motion models, including the unicycle model and the bicycle model to better capture the motion of turning vehicles. Our approach is simple and highly efficient, making it suitable for practical applications in 3D detection systems. We hope that our work will inspire further improvements and applications of the method in the future.

{\small
\bibliographystyle{ieee_fullname}
\bibliography{egbib}
}

\setcounter{table}{0}
\renewcommand{\thetable}{\Alph{table}}
\newpage
\ 
\newpage
\begin{appendix}

\part*{Appendix}

\section{Score Decay Strategy}
We decay the box scores in two circumstances in our frame fusion method.
\paragraph{Weight Decay} The first one is the decayed box confidences which serve as the voting weight in the weighted NMS. For a box $b_i$ from history frame $i$, its decayed confidence will be computed as
\[
    w_i = s_i \cdot d ^{t_i/\tau},
\]
where $s_i$ is the original box confidence score, $d$ is a weight decay factor, $t_i$ is the time lag of history frame $i$ and $\tau$ is the dataset frame interval.
\paragraph{Box Score Decay} The second situation is when a box is solely fused by boxes from history frames, which means it is a new detection box added to the current frame. To avoid affecting the current detection, we decay the scores of these boxes. Two different decay strategies are explored in our experiments. Initially, we directly take the decayed confidence (the voting weight) as the decayed score. The confidence is fused as an attribute in weighed NMS and then serves as the new score. In the experiments, we find that the number of history boxes fused affects the box precision significantly. Thus another decay formula is tried for a fused box $b_f$
\[
    s_f = \frac{d_s\cdot s}{max(N - n_f, 1)},
\]
where $s$ is the original box confidence score, $d_s$ is a score decay factor ($0.6$ in our experiments), $N$ is the frame fusion number and $n_f$ is the number of boxes fused by $b_f$. For the above two strategies, we call them \textbf{decay} and \textbf{divide} strategies respectively.

We construct our score decay strategies from our pioneer works on track-based detection enhancement. They are currently simple and intuitive. We think it is highly feasible to tune or change the strategies for improving our method or apply it in specific scenes.
\section{FrameFusion Configurations}
\begin{table}\footnotesize
  \centering
      \begin{tabular}{@{}lcccc@{}}
    \toprule
    Method     & $d$ & $h_{low}$ & $h_{high}$ & Score Decay\\
    \midrule
    CenterPoint-Based(Waymo) &        0.8&	0.7&	0.7	& decay\\
    CenterPoint-Based(nuScenes) & 0.6&	0.2&	0.7	& decay\\
    Multiple Methods(Table 1) & 0.8 & 0.9	& 0.9	& divide\\
    \bottomrule
\end{tabular}
  \caption{FrameFusion configurations for different experiments. We use the second line for Table 3 and the third line for Table 1. All the other experiments are based on line 1 config.}
  \label{tab:ff_config}
\end{table}
Table \ref{tab:ff_config} presents the parameter configurations for all of our experiments. In our CenterPoint-based experiments, we primarily focus on exploring the performance of the motion models and do not exhaustively tune frame fusion parameters. We set $h_{low}$ to match the original NMS threshold on the dataset ($0.7$ on Waymo~\cite{sun2020scalability}, $0.2$ on nuScenes~\cite{caesar2020nuscenes}), and $h_{high}$ to $0.7$. In follow-up experiments on multiple 3D detection methods to further validate the effectiveness of our frame fusion, we found that setting $h_{low}$ higher consistently produced better fusion results. Two score strategies do not have a clear priority over each other. To sum up, we recommend using $d=0.8,h_{low}=h_{high}=0.9$ on Waymo Dataset and trying both score decay strategies or modifying it if necessary.



\section{Ground-Truth Generation}
For motion parameter regression, we preprocess the dataset with inverse models to generate GT motion parameters before training. Given a pose pair $(p_0, p_{t})$, an inverse model estimates the motion parameter $\hat{\theta_M}$. We introduce our inverse models for unicycle and bicycle models.

For the unicycle model, given $(p_0,p_t)$, the inverse model can be directly inferred from the unicycle forward model:
\begin{equation}
    \label{eq:unicycle_backward}
    \hat{\theta_{U}}=
    \begin{bmatrix}
    V\\
    \omega\\
    \end{bmatrix}=
    \begin{bmatrix}
    \frac{\Delta \phi}{\sin\Delta \phi}(V_x\cos\phi_0+V_y\sin\phi_0)\\
    \Delta \phi / t\\
    \end{bmatrix},
\end{equation}
where $\Delta \phi = \phi_t-\phi_0, V_x=(x_t-x_0)/t,V_y=(y_t-y_0)/t$.

For the bicycle model, we adopt the numerical method to solve $\theta_{B}$. In specific, we use the Gaussian-Newton algorithm to solve a least square problem of pose estimation error. Given a pose pair $(p_0,p_t)$, a motion parameter estimation $\theta_{B}$, the pose estimation error is $r(\theta_{B})=p_t-\hat{p_t}$ where $\hat{p_t}$ is the time $t$ pose estimated with $\theta_{B},p_0$ by the forward model. Thus the target is
\begin{equation}
    \label{eq:bicycle_backward}
    \min_{\theta_{B}} L=\frac{1}{2}r(\theta_{B})^{T}r(\theta_{B}).
\end{equation}

With an initial $\theta_{B}^{0}$, we iteratively update the motion parameter estimation by
\begin{equation}
    \label{eq:bicycle_update}
    \theta_{B}^{s+1}=\theta_{B}^{s}-Inv(J_r) r(\theta_{B}^{s}),
\end{equation}
where $J_r$ is the Jacobian matrix of $r$ about $\theta_{B}$ and $Inv(J_r)$ is the pseudo-inverse of $J_r$. In practice, we terminate the iteration when $L_{k}-L_{k-1}<1e-6$ and output $\theta_{B}^{k}$ as the motion parameter estimation.

$J_r$ can be derived from the bicycle forward model,

\begin{equation}
\label{eq:bicycle_jacobian_backward}
    J_r    =
    \begin{bmatrix}
    t\cos_t & \frac{\partial r_x}{\partial \beta}\\
    t\sin_t & \frac{\partial r_y}{\partial \beta}\\
    \frac{\sin \beta}{l_r}t & \frac{V \cos \beta}{l_r}t\\
    \end{bmatrix},
\end{equation}
\begin{equation*}
\begin{split}
\frac{\partial r_x}{\partial \beta} &= -\frac{l_r}{\sin\beta}[\cot\beta(\sin_t-\sin_0) - (\frac{\partial r_\phi}{\partial \beta}+1)\cos_t+\cos_0] \\
\frac{\partial r_y}{\partial \beta} &= -\frac{l_r}{\sin\beta}[\cot\beta(\cos_t-\cos_0) - (\frac{\partial r_\phi}{\partial \beta}+1)\sin_t+\sin_0],        
\end{split}
\end{equation*}
where $\cos_0=\cos(\phi_{0}+\beta),\sin_0=\sin(\phi_0+\beta),\cos_t=\cos(\hat{\phi_t}+\beta),\sin_t=\sin(\hat{\phi_t}+\beta)$.

For each object at frame $t$, we estimate its motion parameters with object poses in adjacent frames $t-1$ and $t+1$ as $(p_0, p_t)$ in the inverse models.

\section{Frame Fusion Algorithm}
We provide the pseudocode of our proposed frame fusion algorithm (Algo. \ref{algo:frame_fusion}). Readers can refer to it for more algorithm details.

\begin{algorithm}[htbp]
  \caption{Frame Fusion}
  \label{algo:frame_fusion}
  \SetNoFillComment
  \small
  \SetAlgoLined
  \KwData{$D=\{D_{t-i}\}_{i=0}^{N}$: Detections from frame $T-N$ to $T$. $D_{t-i}=\{d_{t-i}\}=\{b_{t-i}, s_{t-i}, c_{t-i},\theta_{t-i}\}$ includes bounding boxes $b_{t-i}$, scores $s_{t-i}$, class labels $c_{t-i}$ and motion parameters $\theta_{t-i}$ of all detection $d_{t-i}$ in frame $T-i$.}
  \KwResult{$D^{f}$: Fused frame $T$.}

  \kwHyper{Target classes $C$, motion model $M$, score decay factor $d$, low and high IoU threshold $h_{low},h_{high}$}
  \For{$i$:$1\rightarrow N$}{
    $t \leftarrow$ Time(Frame $T$) - Time(Frame $T-i$)\;
    $b_{t-i}^{for} \leftarrow$ Forward$_{M}$($b_{t-i}$, $\theta_{t-i}$, $t$)\; \tcp{Forward boxes with motion model $M$} 
    Transform $b_{t-i}^{for}$ to frame T coordinate\;
    $w_{t-i} \gets s_{t-i}\cdot d^{t/\tau}$    \tcp*[l]{Get decayed weight}
    $D_{t-i}^{for} \gets \{b_{t-i}^{for}, s_{t-i}, c_{t-i}, \theta_{t-i}, w_{t-i}\}$\;
  }
  
  $D \gets \cup \{D_{t-i}^{for}\}_{i=0}^{N}$ \tcp*[l]{Dense overlapped frame}
  $D \gets$ Sort($D$, key=$sd$)\;
  $D_{f} \gets \emptyset$\;
  \tcp{Start weighted NMS}
  \For{$c \in C$}{
    $D_{c} \gets \{d|d\in D, d.c=c\}$\;
    $D_{cf} \gets \emptyset$\;
    \tcp{Start fusing boxes with label $c$}
    \For{$d \in D_{c}$}{
    $D_{remove} \gets \{d_r| IoU(d_r.b, d.b)>h_{low}\}$\;
    $D_{merge} \gets \{d_k| IoU(d_m.b, d.b)>h_{high}\}$\;
    \tcp{Select detection subsets with IoU threshold $h_{low},h_{high}$}
    $D_{merge} \gets \{(d.b,d.s,d.\theta,d.w)|d\in D_{merge}\}$\;
    $d_f \gets$ Average($D_{merge}$, weight=$d_m.w$)\;
    \tcp{Box voting in $D_{merge}$ with decayed confidence as weight}
    $d_f.c \gets c$ \;
    $D_c \gets D_c - D_{remove}$\;
    \tcp{Remove overlapped boxes from $D_c$}
    $D_{cf} \gets D_{cf}\cup \{d_f\}$\;
    }
    $D_f \gets D_f \cup D_{cf}$\;
  } 
  \tcp{Decay scores of some boxes}
  \For{$b \in D_f$}{
    \If{$b$ only fused by history boxes}{
        \If{Decay Strategy}{
            $b.s = b.w$\;
        }\ElseIf{Divide Strategy}{
            $b.s = \frac{d_s b.s}{N - n_f}$\;
        }
    }
  }

\end{algorithm}

\section{Implementation Details}
Methods in Table 1 are implemented on OpenPCDet~\cite{openpcdet2020}. We adopt the default configurations for these methods to achieve their reported performances, except for the additional velocity regression. 

Next, we introduce the implementation details for CenterPoint-based motion models, which are based on a 3D detection codebase MMDetection3D \cite{contributors2020mmdetection3d}. We use CenterPoint-Voxel as the backbone. Voxel size is $(0.1,0.1,0.15)$ for Waymo and $(0.075, 0.075, 0.2)$ for nuScenes. We use sweeps $[-4,0]$ for Waymo and $[-10,0]$ for nuScenes as model input. Our models are trained with AdamW optimizer and a cyclic learning rate scheduler with a maximum learning rate 1e-3 and a weight decay $0.01$. Train for 12 epochs on Waymo and 20 epochs on nuScenes. We follow the CenterPoint to use random flipping, global scaling, and ground-truth sampling for data augmentation with the same hyperparameters. For motion parameter regression, we set the loss weight $0.2$ for velocity parameters ($v_x,v_y$ in the constant velocity model and $V$ in the unicycle and bicycle models) and $1.0$ for angular parameters ($\omega$ in the unicycle model, $\beta$ in the bicycle model). For bounding box NMS, we apply circle NMS with $0.7$ threshold on Waymo and $0.2$ threshold on nuScenes. All the models are solely trained and evaluated on the vehicle classes in Waymo and nuScenes Dataset. 

In our experiments, to evaluate the detection result on a certain GT subset such as the turn, straight, and stationary subsets, we first match the original detection to the GT subset and keep the bounding boxes with an IoU over 0.5 with a GT box. The filtered result is then evaluated on the GT subset to test its performance. In the motion state ablation, vehicles with $V<0.1$ are split to stationary. Vehicles with $R<25,V>1$ are split to turn. Others are regarded as straight.

\section{Full Frame Visualization}
\begin{figure*}
    \centering
  \begin{subfigure}[c]{0.48\linewidth}
    \includegraphics[width=\linewidth]{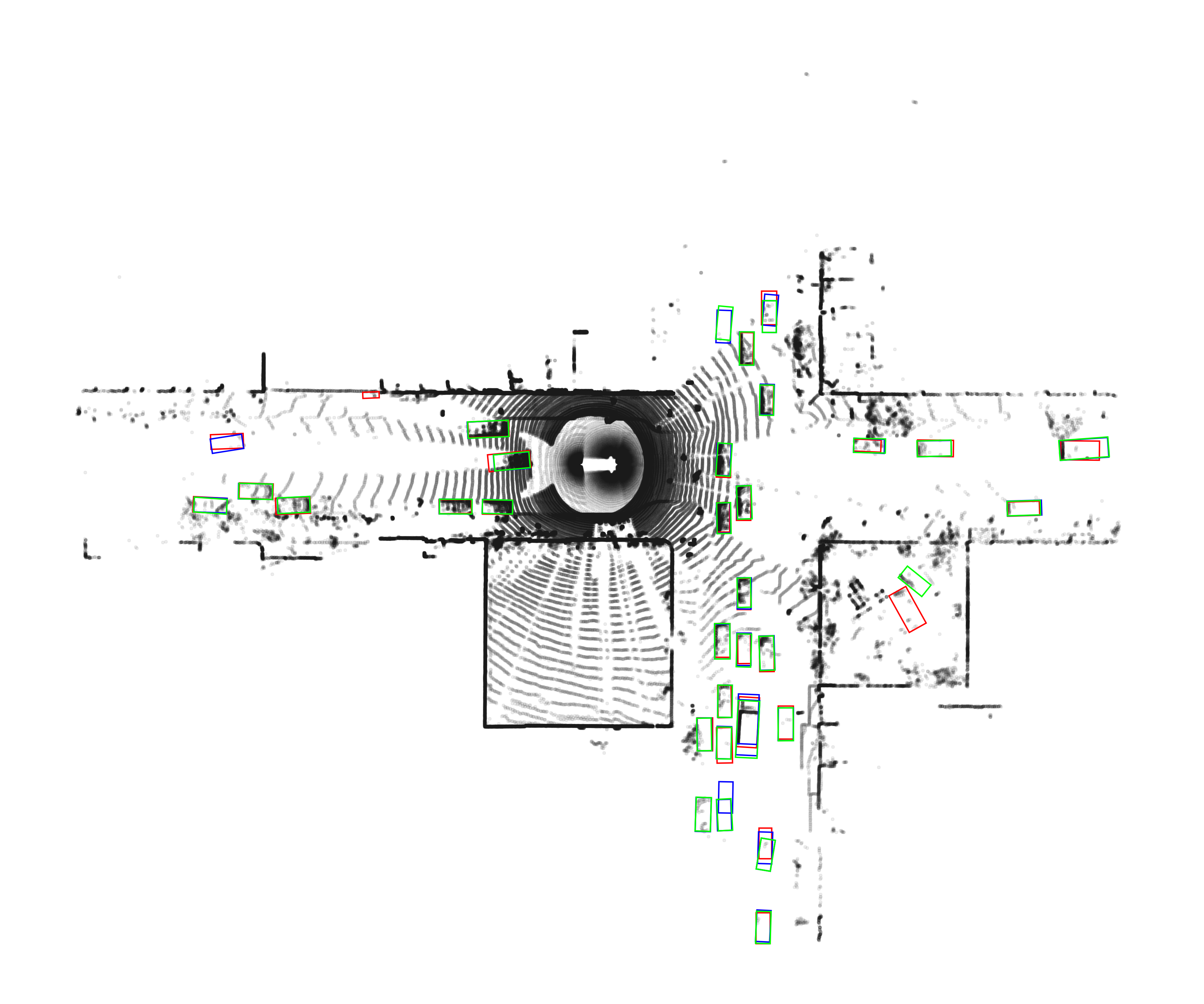}

    \label{fig:lu}
  \end{subfigure}
  \begin{minipage}[c]{0.48\linewidth}
    \centering
\includegraphics[width=\linewidth]{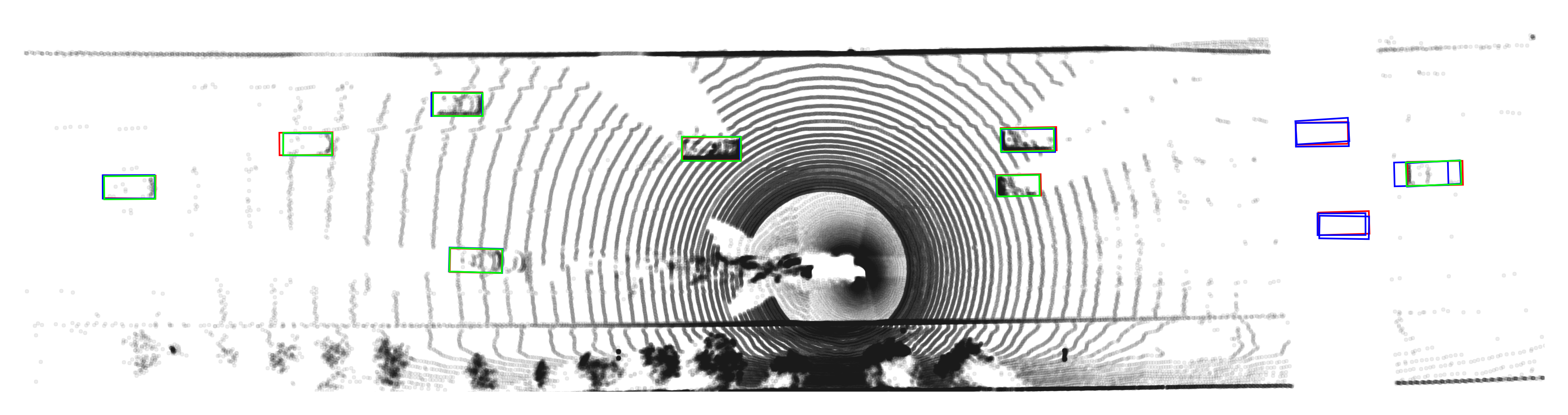}
    \label{fig:ru}
  \end{minipage}
  \\

    \begin{subfigure}[c]{0.48\linewidth}
    \includegraphics[width=\linewidth]{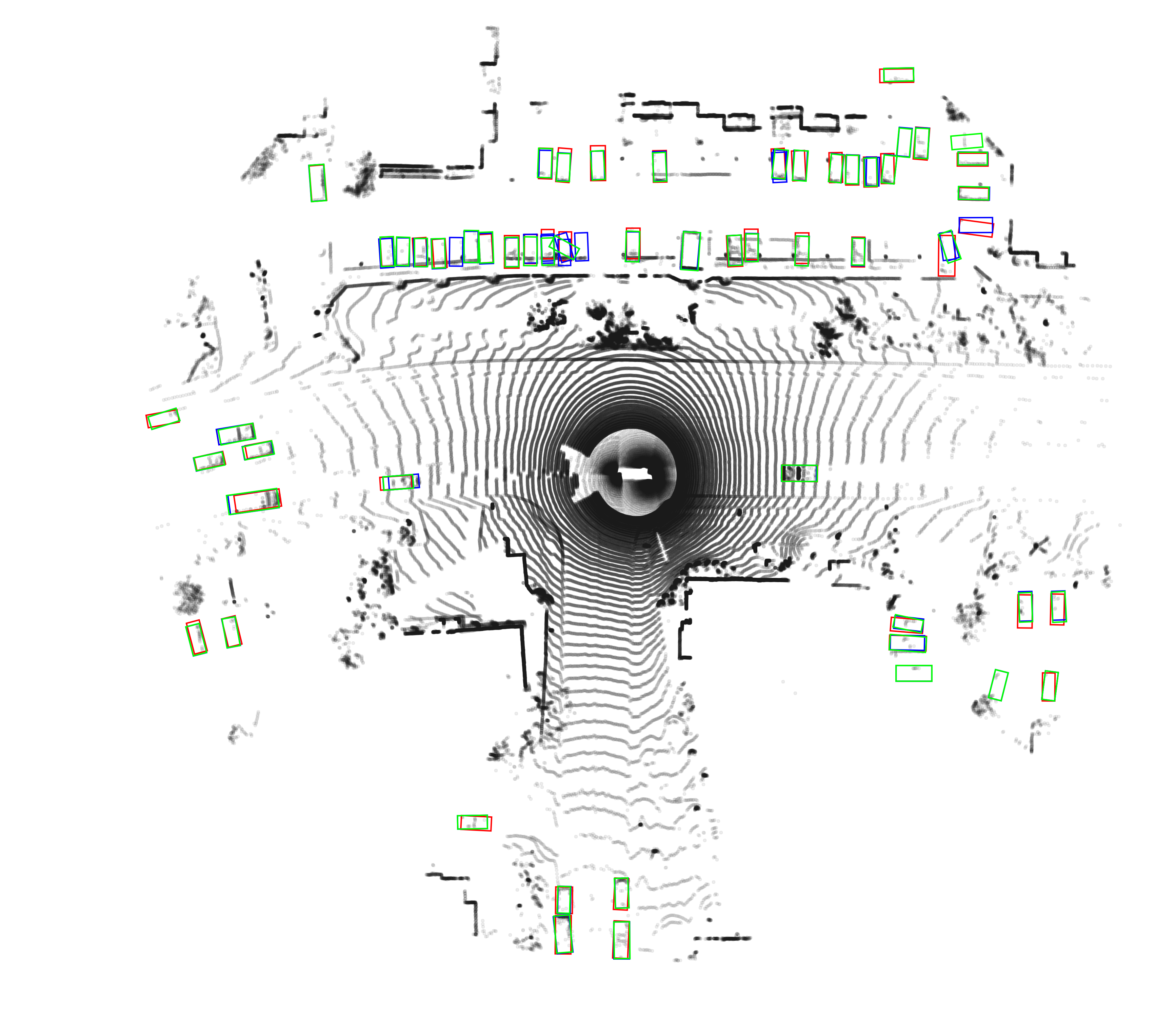}
    \label{fig:lb}
  \end{subfigure}
    \begin{subfigure}[c]{0.48\linewidth}
    \centering
    \includegraphics[width=\linewidth]{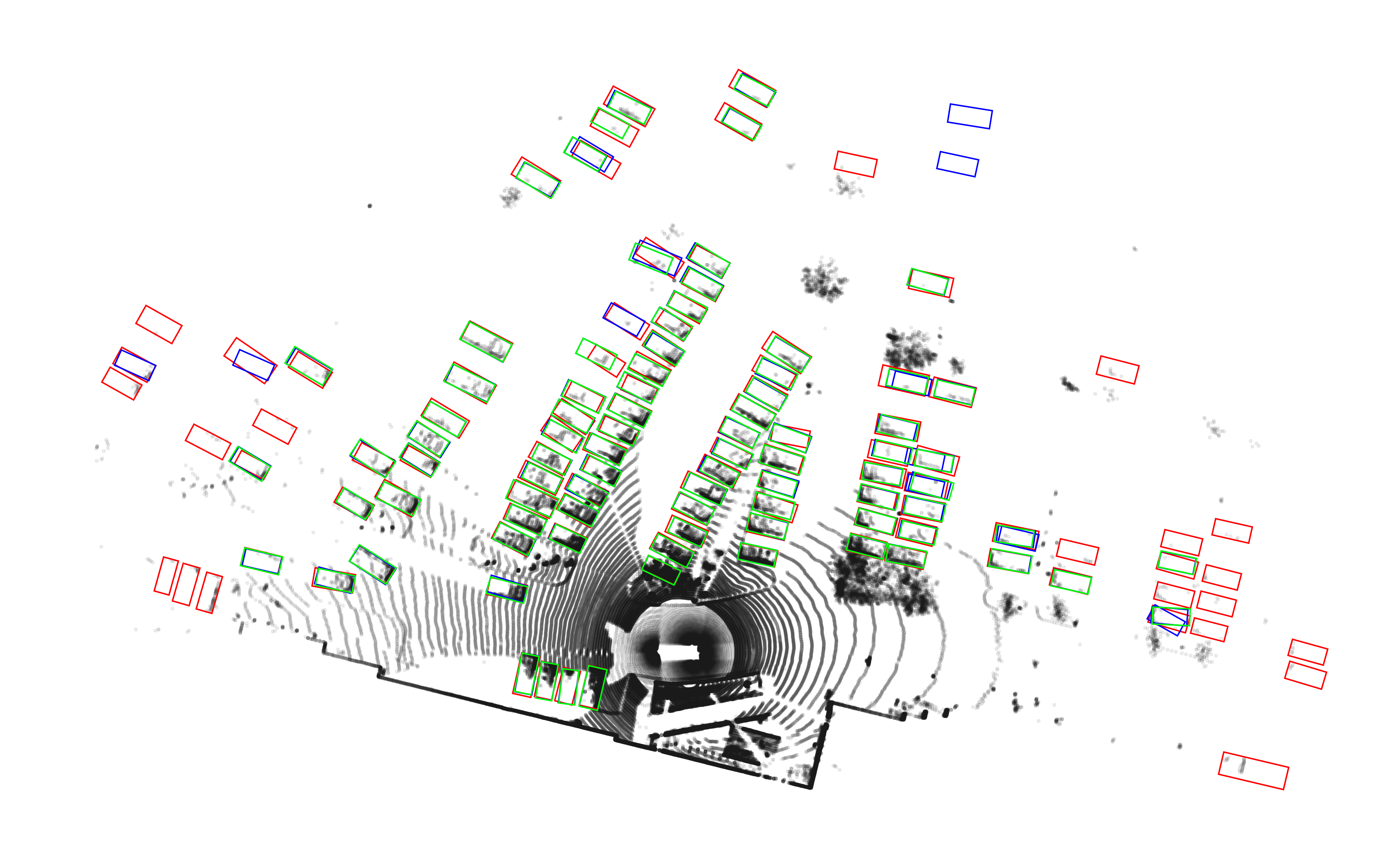}
    \label{fig:rb}
  \end{subfigure}
  \caption{Full frame visualizations for frame fusion result. Red box: GT box. Green box: detection before fusion. Blue box: detection after fusion. Only draw boxes with score $> 0.3$ for better visual effects. Detections unchanged in frame fusion appear as green boxes as we draw green boxes on top of blue boxes. So focus on blue boxes to figure out the frame fusion effect.}
  \label{fig:full_frame_vis}
\end{figure*}

We visualize the frame fusion result in a full frame to demonstrate its effect (Fig. \ref{fig:full_frame_vis}). In the top left figure, we can see most well-detected boxes are unchanged in frame fusion (appear as a single green box). For those missed or inaccurate detections, the frame fusion tries to fix the error and generate new boxes (blue boxes). A few false positives are also produced in this process. The top right figure presents how frame fusion recovers the detection with weak observation. The LiDAR point cloud just loses in a certain range and thus the detector misses the detections in this range. After frame fusion, the boxes are recovered by history detections. The two bottom figures show two complicated scenes with more vehicles.

\end{appendix}
\end{document}